\PassOptionsToPackage{table}{xcolor}
\documentclass[sigconf,nonacm]{acmart}

\AtBeginDocument{}

\usepackage{booktabs}
\usepackage{multirow}
\usepackage{xcolor}
\usepackage{enumitem}
\usepackage{placeins}
\usepackage{array}
\usepackage{etoolbox}
\usepackage{tabularx}

\definecolor{rowgrey}{gray}{0.93}
\AtBeginEnvironment{tabularx}{\rowcolors{2}{rowgrey}{white}}
\AtBeginEnvironment{tabular}{\rowcolors{2}{rowgrey}{white}}

\AtEndPreamble{\theoremstyle{acmplain}\newtheorem{assumption}[theorem]{Assumption}}

\DeclareCaptionType[fileext=lob,placement=tbp]{infobox}[Box][List of Boxes]

\newcommand{\ARzo}{AR$[0,{+}1]$}
\newcommand{\ARpl}{AR$[-5,-1]$}
\newcommand{\Wclean}{$W_{\mathrm{clean}}$}
\newcommand{\Wrelease}{$W_{\mathrm{release}}$}
\newcommand{\Eeff}{E[effect]}
\newcommand{\eightK}{\mbox{8-K}}
\newcommand{\artifactURL}{\url{https://github.com/yebof/track-or-follow}}

\newcommand{\bookSize}{20}
\newcommand{\bandHalfWidth}{10}
\newcommand{\incumbentBuffer}{10}
\newcommand{\capMultiplier}{2}
\newcommand{\capFloor}{0.01}

\newcommand{\nameCapPct}{10}
\newcommand{\industryCapPct}{30}
\newcommand{\turnoverCapPct}{40}
\newcommand{\earningsMult}{0.75}
\newcommand{\earningsSessions}{2}
\newcommand{\sentinelThreshold}{0.8}
\newcommand{\corroborationKeys}{2}
\newcommand{\anchorPriceSd}{2}
\newcommand{\anchorFilingDays}{7}

\newcommand{\universeLookback}{63}
\newcommand{\maxTickers}{3}
\newcommand{\arBetaWindow}{252}

\newcommand{\anchorFilingItems}{1.03, 2.04, 2.06, 3.01, 4.01 and 4.02}
\newcommand{\costBps}{15}

\newcommand{\cutoffTime}{16:00}

\newcommand{\halfLifeWeeks}{52}
\newcommand{\priorSd}{0.04}
\newcommand{\logitClip}{4}
\newcommand{\stdWeeks}{52}
\newcommand{\nTrustedFeatures}{24}
\newcommand{\nUntrustedFeatures}{12}
\newcommand{\newsMaxItems}{40}
\newcommand{\nSummarySentences}{6}
\newcommand{\memoryWeeks}{8}
\newcommand{\nPriorRecords}{5}
\newcommand{\promoWeight}{0.25}
\newcommand{\clusterJaccard}{0.30}
\newcommand{\policyCap}{60}
\newcommand{\nQuestionsNewsItem}{13}
\newcommand{\nEventTypes}{13}
\newcommand{\nEffectLevels}{5}
\newcommand{\nProfileFlags}{21}
\newcommand{\emuModel}{\texttt{gpt-\allowbreak{}4.1-\allowbreak{}mini-\allowbreak{}2025-\allowbreak{}04-\allowbreak{}14}}
\newcommand{\jevModel}{\texttt{jev-\allowbreak{}1.13.0}}
\newcommand{\ngModel}{\texttt{this-\allowbreak{}that-\allowbreak{}model-\allowbreak{}1.0}}

\newcommand{\universeSize}{100}

\newcommand{\nPairedAttacks}{1{,}119}
\newcommand{\nPairedAttacksRegistered}{1{,}400}
\newcommand{\nPropertyCases}{2{,}000}
\newcommand{\hOneMDELow}{0.056}
\newcommand{\hOneMDEHigh}{0.079}
\newcommand{\hOneMDELowRegistered}{0.050}
\newcommand{\hOneMDEHighRegistered}{0.071}
\newcommand{\hTwoMDELow}{0.019}
\newcommand{\hTwoMDEHigh}{0.030}
\newcommand{\hThreeMDELow}{0.021}
\newcommand{\hThreeMDEHigh}{0.043}

\newcommand{\nwLag}{4}
\newcommand{\ciLevel}{95}
\newcommand{\familyAlpha}{0.05}
\newcommand{\bufferWeeks}{4}
\newcommand{\nWeeksRelease}{73}
\newcommand{\priorLookStockWeeks}{600}

\newcommand{\nWeeks}{115}

\newcommand{\hOneASRHolistic}{0.1}
\newcommand{\hOneASRTyped}{10.5}
\newcommand{\hTwoSpearmanEmu}{0.119}
\newcommand{\hThreeDeltaIC}{\ensuremath{-0.016}}
\newcommand{\costPerRebalance}{1.03}

\newcommand{\localizationRate}{88}

\newcommand{\nBoundCases}{8{,}912}
\newcommand{\nBoundViolations}{0}

\newcommand{\latencyPNinetyFive}{1.84}
\newcommand{\mainWallClock}{2.4}

\newcommand{\scalingRequests}{390}
\newcommand{\scalingCost}{0.091}

\newcommand{\maxEntrantRank}{30}
\newcommand{\maxLogitRatio}{0.64}
\newcommand{\maxEntrantWeightRatio}{1.00}

\newcommand{\hOneDelta}{0.104}
\newcommand{\hOneCI}{[0.081, 0.127]}
\newcommand{\hOnePholm}{1.000}

\newcommand{\hTwoCIEmu}{[0.098, 0.140]}

\newcommand{\hTwoPlaceboEmu}{0.046}
\newcommand{\hTwoSpearmanVerb}{0.097}

\newcommand{\hTwoSpearmanRelease}{0.115}
\newcommand{\hThreeDeltaICRelease}{\ensuremath{-0.009}}
\newcommand{\icTextPart}{\ensuremath{-0.006}}

\newcommand{\hThreeCI}{[\ensuremath{-0.029}, \ensuremath{-0.002}]}

\newcommand{\poolLogLossDiff}{\ensuremath{-0.00265}}
\newcommand{\poolLogLossDiffCI}{[\ensuremath{-0.00453}, \ensuremath{-0.00077}]}

\newcommand{\ledgerResidual}{\ensuremath{3\times 10^{-17}}}
\newcommand{\nTrials}{73}
\newcommand{\dsr}{0.41}

\newcommand{\preregDigest}{\NumPlaceholder{preregDigest}}

\newcommand{\hOneSentence}{At the tested attacker and budget, the pre-registered one-sided test that per-item guarded decomposition lowers attack success relative to the holistic reader was not supported (Holm-adjusted $p = \hOnePholm$); the estimate points the other way, with attack success of \hOneASRTyped\% for per-item guarded decomposition and \hOneASRHolistic\% for the holistic reader (difference \hOneDelta{}, \ciLevel\% CI \hOneCI{}); the influence bound capped the damage for every reader, and only per-item readers localized the planted item.}

\newcommand{\hThreeSentence}{No increment was detected at MDE $\approx$ \hThreeMDELow--\hThreeMDEHigh{} IC.}

\newcommand{\hTwoAbstract}{item judgments track announcement-window abnormal returns (weekly Spearman \hTwoSpearmanEmu{})}

\newcommand{\poolSentence}{The learned pool is detectably worse than the equal-weight log pool in log loss (weekly difference, equal-weight minus learned, \poolLogLossDiff{}, \ciLevel\% CI \poolLogLossDiffCI{}).}

\newcommand{\nHTwoUnits}{27{,}556}

\newcommand{\hTwoT}{11.3}
\newcommand{\hTwoCIRelease}{[0.090, 0.140]}

\newcommand{\hTwoPlaceboEmuCI}{[0.023, 0.070]}

\newcommand{\hTwoCIVerb}{[0.054, 0.138]}

\newcommand{\hTwoEmuOnVerb}{0.100}
\newcommand{\hTwoEmuOnVerbCI}{[0.055, 0.142]}

\newcommand{\nVerbUnits}{2{,}000}
\newcommand{\nNGUnits}{500}

\newcommand{\hTwoAUCEarningsEmu}{0.833}
\newcommand{\hTwoAUCEarningsEmuCI}{[0.826, 0.840]}

\newcommand{\hTwoAUCDealEmu}{0.611}

\newcommand{\hTwoAUCMgmtEmu}{0.557}

\newcommand{\dTwoCompanyFull}{3}

\newcommand{\dTwoCannotTell}{94}

\newcommand{\dOneCentreAUC}{0.488}
\newcommand{\dOneCentreP}{0.896}

\newcommand{\dThreePre}{+0.0008}
\newcommand{\dThreePreCI}{[\ensuremath{-0.019}, 0.017]}
\newcommand{\localizationCI}{[80.8, 94.2]}

\newcommand{\asrFamAtwo}{22.5}

\newcommand{\asrFamAfive}{4.0}
\newcommand{\asrFamAsix}{31.5}

\newcommand{\asrFamAtwoCI}{[16.0, 29.3]}
\newcommand{\asrFamAsixCI}{[24.5, 38.6]}

\newcommand{\asrFamAoneN}{173}

\newcommand{\hThreeCIRelease}{[\ensuremath{-0.028}, 0.011]}
\renewcommand{\preregDigest}{\texttt{b8bf8cf6}\ldots\allowbreak\texttt{84b9a}}

\newcommand{\nDeviations}{14}

\newcommand{\feedItems}{100{,}906}

\newcommand{\nItemRowsWindow}{74{,}314}
\newcommand{\nKeptRows}{31{,}848}
\newcommand{\hTwoPholmBound}{\ensuremath{<10^{-6}}}
\newcommand{\hTwoSentenceBound}{Typed item judgments track announcement-window abnormal returns (weekly Spearman \hTwoSpearmanEmu{}, Holm-adjusted $p$ \hTwoPholmBound{}).}
\newcommand{\hTwoMDERealized}{0.031}
\newcommand{\hThreeMDERealized}{0.021}
\newcommand{\mdePowerPct}{80}
\newcommand{\hThreeICPrimary}{\ensuremath{-0.024}}

\newcommand{\hThreeICCodeonly}{\ensuremath{-0.008}}

\newcommand{\hThreeAbstractText}{the confirmatory test detects no rank information over code-only features}

\newcommand{\mOneBRatio}{0.37}
\newcommand{\mOneBPlanted}{0.10}
\newcommand{\alphaSystem}{\ensuremath{-15.7}}

\newcommand{\alphaCodeonly}{\ensuremath{-16.3}}

\newcommand{\dsrHeadline}{0.66}
\newcommand{\nTrialsRun}{71}

\newcommand{\dOneMDE}{0.21}

\newcommand{\dTwoMDECell}{0.112}
\newcommand{\dTwoCompanyFullCI}{[1.0, 8.5]}

\newcommand{\demotionASR}{57}
\newcommand{\demotionCI}{[49, 66]}

\newcommand{\demotionProtected}{64}
\newcommand{\demotionProtectedCI}{[50, 76]}
\newcommand{\demotionProtectedExcl}{0}
\newcommand{\demotionInBand}{46}
\newcommand{\demotionInBandCI}{[32, 60]}

\newcommand{\asrZeroUpper}{2.1}

\newcommand{\voteDelta}{+6.4}
\newcommand{\voteCI}{[4.4, 8.6]}

\newcommand{\nEntryChecks}{7{,}910}
\newcommand{\nProtectionChecks}{1{,}002}
\newcommand{\pTwoProtectedHeld}{290}
\newcommand{\pTwoShareHalved}{74}
\newcommand{\lopoMin}{\ensuremath{-0.007}}
\newcommand{\lopoMinProg}{policy}
\newcommand{\lopoMax}{+0.015}
\newcommand{\lopoMaxProg}{fundamentals}

\newcommand{\wCleanStartProse}{5~July 2024}
\newcommand{\wCleanEndProse}{11~September 2026}

\newcommand{\emuCutoffMonth}{June 2024}
\newcommand{\dOneFirstMonth}{January 2024}
\newcommand{\dOneLastMonth}{June 2025}
\newcommand{\dOneCentreMonth}{July 2024}
\newcommand{\nDOneWindows}{18}
\newcommand{\nSecondaryContrasts}{7}
\newcommand{\nPromotionFamilies}{7}
\newcommand{\nDemotionFamilies}{2}

\newcommand{\IfJevResults}[2]{#1}
\newcommand{\hTwoSpearmanJev}{0.102}
\newcommand{\hTwoCIJev}{[0.017, 0.284]}
\newcommand{\hTwoPlaceboJev}{\ensuremath{-0.024}}
\newcommand{\hTwoAUCEarningsJev}{0.766}
\newcommand{\nJevUnits}{549}
\newcommand{\jevPanelStatus}{descriptive}

\newcommand{\jevFwdWeeks}{3}

\newcommand{\jevFwdSpearmanEmu}{0.076}

\newcommand{\jevFwdPlaceboEmu}{0.058}

\newcommand{\jevAgreeEmuRho}{0.841}

\newcommand{\jevAgreeEmuKappa}{0.77}

\newcommand{\jevUSDPerKItems}{0.18}
\newcommand{\jevLatencyMedian}{0.36}

\newcommand{\jevDTwoCompanyFull}{6.5}

\newcommand{\exPdThreeEpiDeltaIC}{0.0005}
\newcommand{\exPdThreeEpiDeltaICCI}{[\ensuremath{-0.0036}, 0.0046]}

\newcommand{\exPdThreeEpiWeeks}{113}

\newcommand{\exPdThreeAllDeltaIC}{\ensuremath{-0.0026}}
\newcommand{\exPdThreeAllDeltaICCI}{[\ensuremath{-0.0089}, 0.0037]}

\newcommand{\exPdThreeAllWeeks}{98}
\newcommand{\exSubsetWeeks}{60}

\newcommand{\exSubsetICTal}{0.000}
\newcommand{\exSubsetICTalCI}{[\ensuremath{-0.049}, 0.049]}

\newcommand{\exSubsetICTyped}{0.003}
\newcommand{\exSubsetICTypedCI}{[\ensuremath{-0.045}, 0.051]}

\newcommand{\exEoneSlopeHOutlookPosPct}{1}

\newcommand{\exEoneASRHFlipTyped}{21.9}

\newcommand{\exEtwoTraded}{\ensuremath{-0.004}}
\newcommand{\exEtwoTradedCI}{[\ensuremath{-0.019}, 0.011]}

\newcommand{\exEtwoClosedBeforeFillPct}{81}
\newcommand{\exEthreeFlaggedFrom}{2024-11}
\newcommand{\exEthreeFlaggedTo}{2025-02}

\newcommand{\exEfourText}{0.6}
\newcommand{\exEfourTextCI}{[\ensuremath{-10.1}, 11.2]}

\newcommand{\exEfivePriceShare}{7.1}

\newcommand{\exEfiveAuditPrecision}{97}
\newcommand{\exEfiveAuditPrecisionCI}{[92, 99]}
\newcommand{\exEfiveAuditRecall}{65}
\newcommand{\exEfiveAuditRecallCI}{[48, 88]}
\newcommand{\exEfiveAuditMiss}{4}
\newcommand{\exEfiveAuditPrevalence}{10.6}
\newcommand{\exEfiveAuditN}{100}

\newcommand{\exEfiveNoPrice}{0.089}
\newcommand{\exEfiveNoPriceCI}{[0.067, 0.111]}
\newcommand{\exEfivePriceOnly}{0.341}

\newcommand{\exEfivePlaceboNoPrice}{0.039}
\newcommand{\exEfivePlaceboNoPriceCI}{[0.015, 0.062]}

\newcommand{\exEfiveLM}{0.062}

\newcommand{\exEfiveFirmDay}{0.111}
\newcommand{\exEfiveFirmDayCI}{[0.091, 0.131]}
\newcommand{\exEfiveFirmDays}{13{,}287}
\newcommand{\exEfiveConceptEarnings}{0.226}

\newcommand{\exEfiveEitherShare}{9.6}
\newcommand{\exEfiveNoPriceEither}{0.080}
\newcommand{\exEfiveNoPriceEitherCI}{[0.058, 0.103]}

\newcommand{\exEthreePreCutoffWindows}{14}

\newcommand{\exEthreePreCutoffAboveHalf}{13}

\newcommand{\exEsixPubAll}{0.023}
\newcommand{\exEsixPubAllCI}{[0.005, 0.041]}

\newcommand{\exEsixSharePreOpen}{42}
\newcommand{\exEsixPreOpen}{0.105}

\newcommand{\exEsixPreOpenPub}{0.035}

\newcommand{\exEsixShareIntraday}{40}
\newcommand{\exEsixIntraday}{0.155}

\newcommand{\exEsixIntradayPub}{0.020}

\newcommand{\exEsixShareAfter}{17}
\newcommand{\exEsixAfter}{0.060}

\newcommand{\exEsixAfterPub}{\ensuremath{-0.018}}

\newcommand{\exEsixRelNoPrice}{0.081}
\newcommand{\exEsixRelNoPriceCI}{[0.055, 0.107]}

\newcommand{\exEsixPlaceboDiff}{0.073}
\newcommand{\exEsixPlaceboDiffCI}{[0.045, 0.100]}

\newcommand{\exEsevenFinbert}{0.094}

\newcommand{\exEsevenVerbDiff}{\ensuremath{-0.003}}
\newcommand{\exEsevenVerbDiffCI}{[\ensuremath{-0.029}, 0.023]}
\newcommand{\exDOneMinP}{0.005}
\newcommand{\exDOneHolmMinP}{0.09}

\newcommand{\exEeightFinbertPub}{0.004}

\newcommand{\exEeightSecUnits}{2{,}709}
\newcommand{\exEeightSecAfterPct}{72}
\newcommand{\exEeightSecMedianMin}{32}

\newcommand{\exEeightAfterCloseMin}{35}
\newcommand{\exEeightDeskSharePct}{64}
\newcommand{\exEeightDeskPricePct}{0.6}
\newcommand{\exEeightDesk}{0.088}

\newcommand{\exEeightDeskPub}{0.021}

\newcommand{\exEeightNamedSharePct}{36}
\newcommand{\exEeightNamedPricePct}{19.0}
\newcommand{\exEeightNamed}{0.173}

\newcommand{\exEeightNamedPub}{0.032}

\newcommand{\exEeightFramePairs}{53{,}698}
\newcommand{\exEeightFrameKeptPairs}{31{,}848}
\newcommand{\exEeightFrameAll}{0.086}
\newcommand{\exEeightFrameAllCI}{[0.071, 0.101]}
\newcommand{\exEeightFrameKept}{0.113}
\newcommand{\exEeightFrameKeptCI}{[0.092, 0.135]}
\newcommand{\exEeightFrameAllNoPrice}{0.062}
\newcommand{\exEeightFrameAllNoPriceCI}{[0.047, 0.077]}

\newcommand{\exEnineEarnGapMedianMin}{125}

\newcommand{\exEnineEarnRhoPub}{\ensuremath{-0.041}}

\newcommand{\exEnineEarnRhoAccRegular}{\ensuremath{-0.021}}

\newcommand{\exEnineEarnEvRhoAcc}{0.088}
\newcommand{\exEnineEarnEvRhoAccCI}{[0.001, 0.174]}

\newcommand{\litTangNoBaseline}{21}
\newcommand{\litTangEvaluating}{27}

\newcommand{\litLopezDriftSharpeLast}{1.22}

\newcommand{\titleSystem}{Jev-Compatible}
\newcommand{\titleShort}{Track or Follow? A \titleSystem{} Multi-Agent Trading System When Web News Lags the Market}

\begin{document}

\title[\titleShort]{Track or Follow? A \titleSystem{} Multi-Agent Trading System When Web News Lags the Market}

\author{Yebo Feng}
\affiliation{\institution{Nanyang Technological University}\country{Singapore}}
\email{yebo.feng@ntu.edu.sg}
\author{Guangyan Gan}
\affiliation{\institution{Nanyang Technological University}\country{Singapore}}
\email{GUANGYAN001@e.ntu.edu.sg}
\author{Liangrui Pan}
\affiliation{\institution{Nanyang Technological University}\country{Singapore}}
\email{liangrui.pan@ntu.edu.sg}
\author{Ruichao Liang}
\affiliation{\institution{Nanyang Technological University}\country{Singapore}}
\email{ruichao.liang@ntu.edu.sg}
\author{Cong Wu}
\affiliation{\department{School of Cyber Science and Engineering}\institution{Wuhan University}\city{Wuhan}\country{China}}
\email{cnacwu@whu.edu.cn}
\author{Jiahua Xu}
\affiliation{\institution{UCL Centre for Blockchain Technologies and DLT Science Foundation}\city{London}\country{United Kingdom}}
\email{jiahua.xu@ucl.ac.uk}
\author{Yang Liu}
\affiliation{\institution{Nanyang Technological University}\country{Singapore}}
\email{yangliu@ntu.edu.sg}

\begin{abstract}
LLM agents increasingly trade on the Web news they read, yet much Web news about listed firms is published after,
and about, the price move it reports, so a reader can agree with the market by following it instead of reading the
event. We ask whether LLM judgments of Web news track the market's reaction to the reported event or follow it. To
answer, we built a multi-agent trading system whose agents share only point-in-time stores and answer typed questions
through the interface of Jev, a hosted typed decision model; code sets every position under a proved bound on how far
text can move it. In a hash-committed test of a Jev-compatible reader built from a pinned LLM's option probabilities,
\hTwoAbstract{}, also after the LLM's release, and event types agree with the issuers' \eightK{} filings (AUC
\hTwoAUCEarningsEmu{}); Jev's judgments of news published after its own release track the reaction as well
(descriptive). Exploratory analyses show where the association comes from: \exEeightSecAfterPct\% of earnings news items
appear after the issuer's filing, contributor-written items often report the price move, and from the first trade
after publication the association falls to \exEsixPubAll{}. At a weekly horizon, \hThreeAbstractText{}. Under attack,
decision cards traced \localizationRate\% of successful attacks to the planted item. We propose that market-graded tests
of Web-text readers report the criterion from publication, not only from the event.

\end{abstract}



\maketitle

\section{Introduction}\label{sec:intro}

LLM agents increasingly trade on the Web news they read~\cite{xiao2024tradingagents,li2025time}. Whether they
understand that news is a Web-mining question, and news about listed firms makes it testable: the market later records
how each reported event was received. Web news, however, is often published after, and about, the price move it would
be graded against, so a reader can agree with the market by following the move instead of reading the event.

We ask whether LLM judgments of Web news items track the market's reaction to the reported event and the issuer's
filing, or follow them, and what a trading system gains from them. Three confounds stand in the way. The records that
settle the question begin with the event, which can precede the item; the model may already hold those records, since
recall inside its training period cannot be identified from its outputs~\cite{lopezlira2025memorization}; and the item
may itself report the price move. We
analyze when and by whom Web financial news is written and test machine readers of that content against records they
do not write.

We built a multi-agent trading system to answer these questions (\S\ref{sec:system}). Per-item news agents, a policy
agent, and technical, fundamentals and macro agents exchange no messages; they share only write-once point-in-time
stores. All agents answer through one typed interface that follows the schema of Jev, a hosted decision model that
returns one probability per caller-declared option~\cite{sun2026typesafe,ling2026jev}. Because a typed judgment's form
guarantees nothing about its meaning~\cite{sun2026typesafe}, the interface fixes the form and our evaluation tests the
meaning, for any reader that speaks the schema. Code, never a model, sets each position, under a proved
\emph{influence bound} on how far text can move it. The confirmatory reader, \emph{emu}, serves the schema from a
pinned LLM's option probabilities, the label-probability readout that a typed model must beat~\cite{tang2026typed};
Jev answers the same requests on news published after its release. Before the evaluation window was unlocked, we
hash-committed a protocol with three Holm-corrected tests. H2 compares each item's expected effect with the focal
firm's abnormal return over the event session and the next. H3 tests whether the judgments add rank information to
code-only features. H1 tests whether per-item readers with guard questions lower attack success relative to a
\emph{holistic reader} of a stock's items.

We report four findings. \emph{Validity}: the confirmatory test detects a positive association, a weekly Spearman
correlation of \hTwoSpearmanEmu{} (\ciLevel\% CI \hTwoCIEmu{}), and \hTwoSpearmanRelease{} after the LLM's release;
event types agree with the issuers' \eightK{} filings, their mandatory SEC reports of material events.
\IfJevResults{The association also holds on news published after every reader's training: on the \nJevUnits{} items
published after Jev's release, Jev's judgments correlate \hTwoSpearmanJev{} \hTwoCIJev{} with the reaction (placebo
\hTwoPlaceboJev{}; descriptive). }{}\emph{Timing}: exploratory analyses trace the association to how Web news is produced. Of the
earnings items that report an issuer's filing, \exEeightSecAfterPct\% appear after it, and contributor-written items report
the price move far more often than desk items; from the first trade after publication, the association falls to
\exEsixPubAll{} \exEsixPubAllCI{}, for a finance-tuned tone model as well. \emph{Usefulness}: for this learner and
decision layer, the confirmatory test detects no rank information over code-only features at a weekly horizon
(\hThreeDeltaIC{}, CI \hThreeCI{}). \emph{Agents}: every attack stayed within the bound on how far text moves a position, and decision cards traced \localizationRate\% of successful attacks
to the planted item. Planted facts and press releases still moved names inside the band in \asrFamAtwo\% and
\asrFamAsix\% of attacked stock-weeks, and per-item reading alone did not lower attack success relative to a holistic
reader (H1, \hOneASRTyped\% vs.\ \hOneASRHolistic\%): what limits text is the bound, not the reader.

Prior work shows that LLM judgments of news track the initial reaction~\cite{lopezlira2026chatgpt,kirtac2026financial},
and prior tests either grade readers against gold labels~\cite{gilardi2023chatgpt,ibrahim2026evaluating} or collect
forecasts forward in time~\cite{chen2026autonomous,fatouros2026signal}. We contribute:
(i)~a reader-agnostic, point-in-time multi-agent trading system in which the agent decomposition carries the evaluation: per-agent clearances yield an
influence bound that holds for any reader, single-writer stores enforce the look-ahead control, and decision cards
attribute every moved position to an item;
(ii)~a pre-specified, point-in-time test of LLM judgments of Web news against the market's reaction and the issuers'
\eightK{} filings, with placebo, pre-open and post-release estimates\IfJevResults{, including a typed decision model
graded only on news published after its release}{};
(iii)~an exploratory content analysis of Web financial news that separates the association realized after publication
from what precedes it and from items that report the move, and the reporting rule it yields (\S\ref{sec:discussion});
and (iv)~evidence on what validity buys inside a trading system: no rank information at a weekly horizon, consistent
with fast assimilation~\cite{lopezlira2026chatgpt,ke2019predicting}, and a reader that stays steerable by content it
has reason to believe~\cite{hu2026jevadvbench}.

\section{Problem Formulation}\label{sec:setting}

\begin{table}[t]
\caption{Terms used throughout.}
\label{tab:terms}
{\scriptsize\setlength{\tabcolsep}{3pt}
\begin{tabularx}{\linewidth}{p{0.23\linewidth}>{\raggedright\arraybackslash}X}
\toprule
Typed judgment & one probability per declared option, returned without generated text \\
emu; Jev; NG; VERB & confirmatory reader from a pinned LLM's option probabilities; hosted typed model; open typed
model; the same LLM's verbalized probabilities \\
Frame & kept items whose top event type is not stock activity or opinion; H2's units \\
Band & names of trusted rank at most $K+J$, the only names text can admit \\
Signal; system book & selected names at equal weight; names selected with exclusions and an incumbent buffer, sized under caps and risk rules \\
C; R; X; D & confirmatory; registered; exploratory (not pre-registered); descriptive \\
\bottomrule
\end{tabularx}}
\end{table}

\paragraph{Readers}
A \emph{closed-choice reader} receives a state and a question with declared options and returns one probability per
option; we call that distribution, returned without generated text, a \emph{typed judgment}. All our readers sit behind
one interface that follows the documented schema of Jev (\jevModel{}) and answer the same frozen bank of typed
questions. The confirmatory reader, emu, serves the schema from a pinned generative LLM, \emuModel{},
by renormalizing its option-letter probabilities. The \emph{verbalized
contrast} (VERB) asks the same LLM to state its probabilities in text, so its judgments have the same form but are not typed
judgments. \emph{NG} is an open non-generative typed model, \ngModel{}, which reads each answer from its own slot in
one forward pass~\cite{cheng2026thisthat}. Jev answers through the same interface, served by
a third-party inference host (Cloudflare Workers AI).

\paragraph{Validity criterion}
We call a judgment valid if what its options say about an event agrees with records the reader does not write: the
market's reaction to the event the item reports, and the issuer's own filing. The effect question asks how the
development changes the focal firm's business prospects, on five levels from ``clearly harmful'' through ``neutral or
mixed'' to ``clearly helpful''; \Eeff{} is the probability-weighted level. The announcement-window criterion \ARzo{} is
the focal firm's return minus a trailing \arBetaWindow{}-session point-in-time beta times the S\&P~500 ETF's return,
over the \emph{event session} (the item's session if it appears before the close, else the next) and the session after
it. The H2 statistic is the weekly Spearman correlation of \Eeff{} with \ARzo{}, averaged over weeks and tested
one-sided with Newey--West standard errors (\nwLag{} lags). Event types are compared with the \eightK{} item codes the
firm files from two days before to one day after the item. Both windows are anchored on the event, not on publication:
part of \ARzo{} can precede an item that appears during the session, and a filing can precede the item that reports it
(Fig.~\ref{fig:horizon}a). The test therefore measures criterion validity, and two further measurements locate the
association in time: a pre-open control and an exploratory \emph{publication-aligned} criterion, measured from the first
trade after publication.
Of the \nItemRowsWindow{} item--firm pairs (\emph{items} from here on) that reached the per-item questions, emu's
answers kept \nKeptRows{}: about the focal firm, a new development, and not quarantined. The \nHTwoUnits{} kept items
whose top event type is not ``stock activity or opinion'' form the H2 \emph{frame}, over \nWeeks{} weeks;
App.~\ref{app:validity} repeats it on frames the reader does not select.

\paragraph{Look-ahead}
Later outcomes can reach a judgment through recall in the model's weights, identity cues in the state and, specific
to Web news, the item itself, which may report the move. Our controls, in order of strength, are the weeks after the
LLM's release~\cite{paleka2026pitfalls}, a window that starts after the documented cutoff and a buffer of
\bufferWeeks{} decisions, blinded states with measured re-identification, and a date-only probe.

\section{Design of the Multi-Agent Trading System}\label{sec:system}

\begin{figure}[t]
  \centering
  \includegraphics[width=\columnwidth]{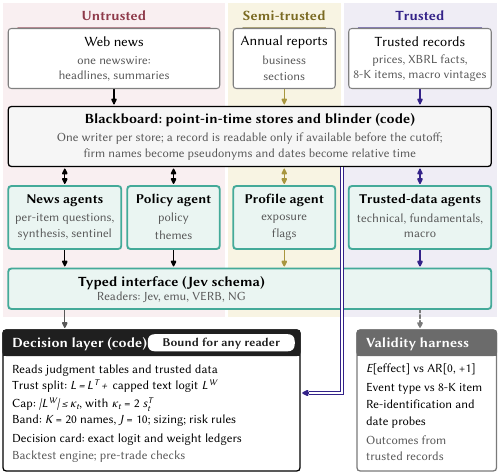}
  \caption{The multi-agent trading system. Agents share only point-in-time stores and answer through one interface that
    follows Jev's schema; code decides under a bound that holds for any reader, and a validity harness compares the
    judgments with records the reader does not write.}
  \Description{Diagram with three vertical trust lanes: untrusted Web news, semi-trusted annual-report text and trusted
    prices, filings and macroeconomic data. Each lane feeds a shared blackboard of point-in-time stores with a blinder.
    Below it sit the agents: news agents and a policy agent in the untrusted lane, a profile agent in the semi-trusted
    lane, and technical, fundamentals and macro agents in the trusted lane; each reads from and writes to the
    blackboard and answers through one typed interface that follows the Jev schema (readers Jev, emu, VERB, NG). Below, a
    decision layer in code (trust split, capped text logit, band, sizing, risk rules, decision card) reads the
    judgments and, through a separate double line that crosses the interface bar, trusted data; a tag says the
    bound holds for any reader. A validity harness, marked evaluation only, compares judgments with abnormal returns,
    8-K items and re-identification and date probes, with nothing flowing back.}
  \label{fig:testbed}
\end{figure}

Four design decisions make the system an instrument for testing readers of Web text rather than a pipeline around
one model: agents ask \emph{typed question programs} instead of exchanging free text; they share only
\emph{point-in-time stores} with single writers and trust labels; a \emph{trust-partitioned learner} feeds a
\emph{band-guarded decision layer} whose influence bound holds for any reader; and reading is separated from deciding,
so every reader is re-evaluated on identical decisions.

\subsection{Architecture and Weekly Cycle}\label{sec:sys-arch}
The system allocates a weekly long-only portfolio over the \universeSize{} most traded S\&P~500 members, without
leverage or short positions and with name, industry and turnover caps. We chose trading because its outcomes are
recorded, public and time-stamped, so every judgment the system acts on can later be checked. Each agent is a question
program with its own state, clearance and store rights. Agents exchange no messages and coordinate only through
write-once stores, a blackboard architecture~\cite{hayesroth1985blackboard} (Fig.~\ref{fig:testbed}). No model output
names an action or a quantity; language models sit upstream of calibration, risk and execution~\cite{ye2026alpha}.
Each decision runs six steps. Code freezes a point-in-time context at the cutoff $c_t$ (\cutoffTime{}~ET on the week's
last session): the universe, daily bars up to that session, news items created and last updated before $c_t$, XBRL
facts filed before the decision day, macroeconomic series by vintage, and store records available before $c_t$. The
macro and policy agents run once for the market; the technical, fundamentals and news agents run for every name in
parallel; labels that matured before $c_t$ join the training set and the learner is refit; the decision layer selects
and sizes the book and writes decision cards; and orders fill at the next open. Only the news agent's novelty check and
weekly synthesis read earlier weeks, so a backtest runs all other requests of all weeks in parallel. The weekly
clock is deliberate: it leaves time to read every item of every name before the decision, it is the horizon at which
a judgment of business prospects should matter, and a cutoff at the close with a fill at the next open keeps every
input strictly before the trade.

\subsection{Typed Question Programs}\label{sec:sys-typed}
A request holds a \emph{state}, a JSON object built in code, and all questions of one stage, each with declared
options: a yes/no question, a choice, or a score on ordered levels. The reader returns one probability per option.
We ask typed questions instead of exchanging generated text for three reasons. A probability over declared options is
a feature that a classical learner can weigh and that can be graded against a later record; no model output can carry
an instruction to the next component, because there is no free text to pass on; and any reader that speaks the schema,
a hosted typed model, an LLM's option probabilities or an open model, plugs into the same slot.
Each decision is asked in exactly one phrasing, and code never derives one answer from another, so every probability in
the system has one source that can be audited. Only code turns answers into features: the log-odds of a yes/no answer
clipped at $\pm$\logitClip{}, the value-weighted mean of an ordered choice, or the centred expected level of a score;
an abstention sets the feature to zero, so it contributes nothing to the logit. States are blinded
by construction: every universe firm becomes \texttt{Company-}$h$, with $h$ the first hex digits of a salted SHA-256
of its symbol, and its names, past tickers and aliases are replaced, with aliases mined only from items published before
$c_t$; dates become relative times, and numbers appear as buckets or percentiles, never as dollar amounts, share prices
or index levels. One interface adds a response cache, a spend ledger with hard caps and rate limits, and stores every
request with its state hash, model pin and question-bank hash.

\subsection{News Agent: From Items to Typed Events}\label{sec:sys-news}
The news agent reads, per stock-week, the \newsMaxItems{} most recent items that tag the firm and at most \maxTickers{}
tickers, after sanitation and exact-headline deduplication. A screening stage asks of the headline and each of up to
\nSummarySentences{} summary sentences whether it reports a development and whether it judges one, and keeps
reporting sentences that do not judge; an item with nothing left is quarantined. Screening removes the opinion,
promotion and instructions that Web text mixes with reports before any judgment is asked. A per-item stage asks
\nQuestionsNewsItem{} questions in one request: relevance to the focal firm, novelty, the event type (\nEventTypes{}
options), the effect on business prospects on \nEffectLevels{} levels from ``clearly harmful'' to ``clearly helpful'',
importance, whether reported figures beat expectations, whether the event was scheduled or is unconfirmed, a
serious-problem question with its kind and status, and two guards. An item is kept if relevance and novelty both reach
one half; a ``yes'' to the instruction guard quarantines it, and to the promotion guard sets its weight $\rho$ to
\promoWeight. Asking per item, not per stock, makes each judgment a testable unit with its own publication time,
the unit that H2 grades and that a decision card can name.
Code then clusters kept items of the same top type whose blinded headlines overlap (Jaccard similarity at least
\clusterJaccard), and asks whether each cluster $e$ repeats one of up to \nPriorRecords{} event-store records of the
last \memoryWeeks{} weeks, giving novelty $\nu_e=1-\max_jP(\text{same as record }j)$. A planted claim repeated across
weeks is thus discounted by the reader's own answers. With $d_e$ the centred expected effect, $m_e$ the importance and
$u_e$ the probability that the event is unconfirmed, the news impact of stock $i$ is
\begin{equation*}
f_i=\operatorname{clip}_{3}\Bigl(\sum_{e}\operatorname{clip}_{3/2}\bigl(\rho_e\,\nu_e\,m_e\,d_e\,(1-u_e/2)(1+\ln w_e/4)\bigr)\Bigr),
\end{equation*}
where $w_e$ is the cluster's width and $\operatorname{clip}_c$ clips at $\pm c$. Clustering counts a syndicated story
once, the logarithm rewards corroboration with diminishing returns, and the two clips keep any single cluster or any
flood of items from dominating a stock's news signal. Kept, non-promotional clusters are written to the event store with their provenance, and a repeat
supersedes its record. A weekly synthesis question reads code-rendered labels of this week's and earlier events and the
stock's trusted market-relative reaction, so no component that synthesizes an answer reads raw
text~\cite{yu2026cordon}. The highest serious-problem probability among a stock's kept items is the \emph{sentinel}.

\subsection{Other Agents over Point-in-Time Stores}\label{sec:sys-stores}
The technical agent reads bucketed price action and market context and asks about trend, pullbacks, capitulation,
breakouts and overextension. The fundamentals agent reads growth, margins, leverage and valuation as percentiles within
the universe and the industry, and asks about growth stage, profitability, financial strain and valuation. The profile
agent reads each annual report's business section once and writes \nProfileFlags{} exposure flags (customers, trade,
regulation, macro, technology) to the filing-profile store; this issuer-written text is \emph{semi-trusted} and read
only for flags. The policy agent reads up to \policyCap{} policy items a week, drawn by a fixed hash, and asks for
theme, stage, effect on profit conditions, uncertainty and affected industries; code maps active themes to stocks
through the profile flags. The macro agent
reads time-blinded indicator percentiles and the scheduled rate-decision calendar and asks about regime, growth,
inflation, policy direction, stress and per-industry tailwinds.
Three stores carry information across weeks: a weekly blackboard, the event store and the filing-profile store. Each
has one writer, its records are immutable and time-stamped, and each record carries the taint of its source (trusted,
semi-trusted or untrusted Web text), never lowered. Agent $a$ at decision $t$ reads exactly
\begin{equation*}
\mathcal{R}(a,t)=\{r:\ \tau_{\mathrm{avail}}(r)<c_t,\ \mathrm{writer}(r)\in\mathcal{A}(a),\ \mathrm{taint}(r)\le\mathrm{clear}(a)\},
\end{equation*}
so look-ahead through a store and untrusted text in a trusted agent's state are excluded by construction, not by
convention. Single writers make every record's provenance unambiguous, and immutability means that a later week can add
a record but never rewrite what an earlier decision read. Annual-report text is semi-trusted because the issuer writes
it and it changes slowly; reading it only for exposure flags lets policy news reach the stocks it concerns without
letting issuer text move a position directly.

\subsection{Trust-Partitioned Learning and the Band-Guarded Decision Layer}\label{sec:sys-decision}
One online logistic model predicts whether a name's open-to-open return over the holding week beats the universe
median. Its \nTrustedFeatures{} trusted features, code factors and the technical, fundamentals and macro judgments, form
the trusted logit $L^T$; its \nUntrustedFeatures{} text features, from the news agent, the sentinel and the policy
agent, form the text logit $L^W$, fitted with $L^T$ as an offset so that text explains only what trusted data leaves
unexplained. Features are standardized over the trailing \stdWeeks{} weeks; slopes have Gaussian priors centred on zero
with standard deviation \priorSd{}, so a feature moves the logit only as far as the data support it; observations decay with a
half-life of \halfLifeWeeks{} weeks; and the model is refit from its prior at every decision. Fitting text on what
trusted data leaves, rather than in one pooled model, is what makes text's influence measurable and boundable: the
trusted logit, and with it the band and the trusted targets, cannot change under any attack on text, and skeptical
priors keep a learner fitted on a short weekly history from chasing noise. The decision logit is
\begin{equation*}
L_i=L^T_i+\operatorname{clip}_{\kappa_t}\bigl(L^W_i\bigr),\qquad \kappa_t=\mu\,s^T_t,
\end{equation*}
with $s^T_t$ the cross-sectional standard deviation of $L^T$ and $\mu={}$\capMultiplier{}. Whatever the untrusted items
say and whichever model reads them, text can then (i)~admit a name only from the \emph{band}, the names of trusted rank
at most $K+J$ ($K={}$\bookSize{}, $J={}$\bandHalfWidth{}); (ii)~not raise a weight above its trusted target;
(iii)~not remove a protected holding (trusted rank at most $K-J$) without a trusted anchor (a severe \eightK{} item or
an abnormal price move); and (iv)~move a name's logit by at most $2\kappa_t$, in this week or through any earlier week's
poisoned records. The band admits text where it is useful, in ranking names that trusted data already rates near the cut, and denies it
where it is dangerous, in promoting names that trusted data rates low. The \emph{signal book} holds the selected names
at equal weight. The \emph{system book} also applies exclusions and an incumbent buffer, and sizes names by
inverse volatility within caps of \nameCapPct\% per name, \industryCapPct\% per industry and \turnoverCapPct\%
turnover, scales a name with earnings due within \earningsSessions{} sessions by \earningsMult{}, halves a name whose
sentinel reaches \sentinelThreshold{}, and excludes it only if \corroborationKeys{} distinct bylines report the problem
and a trusted anchor fires. App.~\ref{app:proof} proves the bound relative to trusted targets, for an attacker who
controls text but not prices.

\subsection{Decision Cards and Paired Re-evaluation}\label{sec:sys-cards}
Because $L$ is additive in slope-weighted features up to the clip, each held, entering or exiting name gets a
decision card whose logit and weight ledgers sum exactly to the decision and assign every contribution to an agent and,
for news, to an item; a moved position therefore traces to the item that moved it (\S\ref{sec:res-planted}), which a
holistic reader cannot provide. Reading and deciding are separate stages that share only judgment tables, in the
capture-once pattern of~\cite{cai2026openpm}: the decision layer, learner variants and ablations are re-run on frozen
judgments at no model cost, and any reader that speaks Jev's schema is re-evaluated on the same weeks, names, band and
cap, so every comparison in \S\ref{sec:results} is paired. The decomposition thus provides what the evaluation needs:
a bound for any reader, a look-ahead control for states, per-item judgments that H2 tests against the market, and
attribution.

\subsection{Implementation and Scale}\label{sec:sys-impl}
On the confirmatory reader, a full run over the evaluation window, all agents, names and weeks, takes \mainWallClock{} hours of wall-clock time, nineteen in twenty requests return within \latencyPNinetyFive{}~s, and reading costs \$\costPerRebalance{} per weekly rebalance at the pinned
model's list price. Cost grows linearly with coverage: each further ten names add \scalingRequests{} requests and
\$\scalingCost{} a week, because stock-level requests concern one name and its items. The bound is tested twice: property tests assert each of its parts on at least
\nPropertyCases{} randomized cases with arbitrary untrusted inputs, and on the real runs a worst-case reader, which sets
the text logit of each name outside the band to its upper cap and all others to the lower cap and drives every
protected name's problem signals to their maximum, produced \nBoundViolations{} violations in
\nBoundCases{} checks (App.~\ref{app:proof}).

\section{Experiments}\label{sec:results}

\subsection{Setup and Pre-Specification}\label{sec:protocol}

\paragraph{Data and windows}
The universe holds the \universeSize{} S\&P~500 members with the largest median dollar volume over the
\universeLookback{} sessions before each month, re-ranked monthly from dated constituent lists. News comes from
Benzinga, a commercial newswire syndicated on the Web, delivered through Alpaca's news API (headline and summary), \feedItems{} items
in the evaluation window. Publication time is the vendor record's creation time, checked against SEC
acceptance times in \S\ref{sec:res-validity}. \Wclean{} holds the \nWeeks{} weekly decisions from \wCleanStartProse{}
to \wCleanEndProse{}, after the \emuCutoffMonth{} cutoff documented for emu's LLM and a buffer; \Wrelease{} holds the
\nWeeksRelease{} decisions after that LLM's release (Table~\ref{tab:models}). Trades cost \costBps{}~bps per side. Emu
answers every request; the verbalized contrast and NG answer the news requests of \nVerbUnits{} and \nNGUnits{} frame
items drawn by a seeded rule stratified by year and event type, scored by the Spearman correlation pooled over items.

\paragraph{Pre-specification and status words}
Before the window was unlocked, we hash-committed the question bank, prompts, model pins, decision layer, attack
templates, seeds and analysis code; a lock in the analysis code blocked every outcome-dependent computation until the
pre-registration's digest (SHA-256 \preregDigest{}) was recorded, and the document is in the code repository
(\artifactURL). \emph{Confirmatory}: H1--H3 on emu, one-sided, Holm-corrected at \familyAlpha{}, with result
sentences fixed in advance. \emph{Registered}: everything else fixed in advance, with \ciLevel\% intervals; the
\nSecondaryContrasts{} secondary contrasts of H1 are Holm-corrected among themselves. An unmarked number is a registered
estimate. \emph{Exploratory (not pre-registered)}: specified after the unlock and always labelled.\IfJevResults{
\emph{Descriptive}: the Jev rows.}{} H2's minimum detectable effect is \hTwoMDELow--\hTwoMDEHigh{} at \mdePowerPct\%
power; the \nDeviations{} logged deviations touch no confirmatory statistic (App.~\ref{app:prereg}).

\paragraph{Probes, usefulness and attacks}
A date-only probe (D1) and a re-identification probe (D2) bound look-ahead (App.~\ref{app:validity}). H3 compares the
rank information coefficient (IC), the weekly Spearman correlation of the combined logit with the holding week's
relative return, with that of a \emph{code-only twin} without typed judgments. The attacker controls any number of
news items under any bylines and knows the architecture and the week's coefficients, but cannot write trusted inputs.
The \nPromotionFamilies{} promotion families succeed if the attacked name, in the band but outside both unattacked
books, enters the signal book; the \nDemotionFamilies{} demotion families succeed if a held name is removed or halved.
H1 compares the \emph{per-item reader} (\S\ref{sec:sys-news}) with a \emph{holistic reader} that asks one typed
question per stock-week over the same items, in the same slot, band and cap, each with its own learned text slope.

\begin{table*}[t]
\caption{Validity of item judgments against the market's reaction and the issuer's \eightK{} filings: (A) the
confirmatory reader, (B) the verbalized contrast. \hTwoSentenceBound{} Newey--West $t={}$\hTwoT{}.}
\label{tab:validity}
{\footnotesize
\setlength{\tabcolsep}{5pt}
\begin{minipage}[t]{0.515\linewidth}
\begin{tabularx}{\linewidth}[t]{c>{\raggedright\arraybackslash}Xrl}
\toprule
\multicolumn{4}{l}{\textit{A. Confirmatory reader (emu), all 27{,}556 frame items}} \\
\midrule
& Measure & Est. & 95\% CI \\
\midrule
C & \ARzo{}, weekly mean (115 weeks) & 0.119 & [0.098, 0.140] \\
R & \quad after the model's release (73 weeks) & 0.115 & [0.090, 0.140] \\
R & \quad published before 09:30 ET & 0.101 & [0.079, 0.124] \\
R & \quad declared prior looks excluded & 0.119 & [0.098, 0.140] \\
R & \quad first 13 weeks dropped & 0.115 & [0.094, 0.137] \\
R & Placebo \ARpl{}, weekly mean & 0.046 & [0.023, 0.070] \\
R & \ARzo{}, pooled over items & 0.122 & [0.104, 0.139] \\
\midrule
\multicolumn{4}{l}{\textit{8-K agreement (AUC) and calibration}} \\
R & Earnings results vs Item 2.02 (7{,}258 pos.) & 0.833 & [0.826, 0.840] \\
R & Deal or investment vs Items 1.01/2.01 (330 pos.) & 0.611 & [0.550, 0.668] \\
R & Management or labor vs Item 5.02 (1{,}144 pos.) & 0.557 & [0.530, 0.586] \\
R & Beat-expectations judgment vs sign of \ARzo{} ($n$ 2{,}435) & 0.605 & --- \\
R & ECE, earnings vs Item 2.02 & 0.174 & --- \\
\bottomrule
\end{tabularx}
\end{minipage}
\hfill
\begin{minipage}[t]{0.465\linewidth}
\begin{tabularx}{\linewidth}[t]{>{\raggedright\arraybackslash}Xrl}
\toprule
\multicolumn{3}{l}{\textit{B. Verbalized contrast (VERB) on its sampled frame items}} \\
\midrule
Measure (pooled over items) & VERB & 95\% CI \\
\midrule
\ARzo{} & 0.097 & [0.054, 0.138] \\
\quad emu on the same items & 0.100 & [0.055, 0.142] \\
Placebo \ARpl{} & 0.048 & [0.002, 0.094] \\
Agreement with emu, $\rho$ of \Eeff{} & 0.81 & --- \\
Agreement with emu, $\kappa$ of top type & 0.73 & --- \\
Earnings vs Item 2.02 (532 pos.) & 0.718 & [0.694, 0.741] \\
Deal vs Items 1.01/2.01 (25 pos.) & 0.587 & [0.510, 0.688] \\
Management vs Item 5.02 (93 pos.) & 0.583 & [0.543, 0.630] \\
Beat-expectations vs sign ($n$ 490) & 0.560 & --- \\
ECE, earnings vs Item 2.02 & 0.161 & --- \\
Sampled items (eligible by own answers) & 2{,}000 & (1{,}729) \\
\bottomrule
\end{tabularx}
\end{minipage}
\par\smallskip
\begin{minipage}{\linewidth}\scriptsize\raggedright
C: confirmatory (one-sided, Holm over H1--H3); R: registered estimate, 95\% interval, no test; ---: not computed; pos.: positives. VERB: the same LLM's verbalized probabilities on a seeded sample of frame items, pooled over items. NG: Table~\ref{tab:ng}.
\end{minipage}
}

\end{table*}

\begin{figure*}[t]
  \centering
  \includegraphics[width=\textwidth]{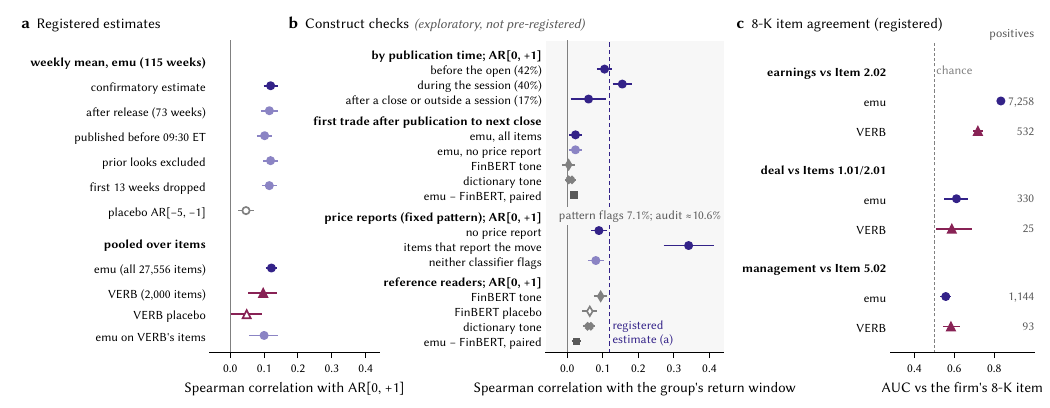}
  \caption{Judgments track the reaction to the reported event, but mostly a reaction that precedes the item. (a)
    Registered estimates, (b) exploratory construct checks, (c) registered \eightK{} agreement. Open markers: placebo
    windows; light indigo: emu's sensitivities; grey: reference readers. \ciLevel\% intervals; intervals narrower than
    the marker are hidden.}
  \Description{Three panels of horizontal interval plots. Panel a lists the confirmatory estimate with its registered
    sensitivities, all of similar size, the smaller placebo, and pooled values for the confirmatory reader and
    the verbalized contrast with its placebo. Panel b, shaded and labelled exploratory, shows the statistic by
    publication time (largest for items published during the session); from the first trade after publication it is
    near zero for emu and zero for two reference readers, with a small positive paired difference between emu and
    FinBERT; it is lower without price-report items and far higher on items that report the move; the reference readers
    lie below the registered estimate on the announcement window. Panel c shows AUCs against 8-K items: high for
    earnings results versus Item 2.02, lower for deals and management changes.}
  \label{fig:validity}
\end{figure*}

\subsection{Typed Judgments Track the Reaction to the Reported Event}\label{sec:res-validity}

The confirmatory test supports H2 (Table~\ref{tab:validity}A, Fig.~\ref{fig:validity}a). Emu's item judgments rank the
focal firm's announcement-window reaction, with a weekly Spearman correlation of \hTwoSpearmanEmu{} (\ciLevel\% CI
\hTwoCIEmu{}; MDE at the realized standard error \hTwoMDERealized), and no registered control or sensitivity removes it. The
placebo \ARpl{} is positive and smaller, \hTwoPlaceboEmu{} \hTwoPlaceboEmuCI{} (exploratory paired difference
\exEsixPlaceboDiff{} \exEsixPlaceboDiffCI{}). The association also holds on frames that the reader does not
select (App.~\ref{app:validity}).
On its \nVerbUnits{} items, the verbalized contrast gives \hTwoSpearmanVerb{} \hTwoCIVerb{} and emu \hTwoEmuOnVerb{}
\hTwoEmuOnVerbCI{} (exploratory paired difference \exEsevenVerbDiff{} \exEsevenVerbDiffCI{}). NG's rows, on \nNGUnits{} items, are in the appendix (Table~\ref{tab:ng}).

Event-type judgments agree with the issuers' filings (Fig.~\ref{fig:validity}c). The probability of ``earnings
results'' ranks items whose firm files an Item~2.02 \eightK{} around the item at AUC \hTwoAUCEarningsEmu{}
\hTwoAUCEarningsEmuCI{}.
Agreement is weaker for deals (\hTwoAUCDealEmu{}) and management changes
(\hTwoAUCMgmtEmu{}), whose item codes are coarse legal categories~\cite{dolphin2026grounded}.

\begin{table}[t]
\caption{Who writes Web financial news and when it appears (exploratory, except the registered all-items \ARzo{}
value): share of frame items, share that reports the price move, and the weekly Spearman correlation $\rho$ of
\Eeff{} with \ARzo{} and with the return from the first trade after publication.}
\label{tab:bylines}
{\scriptsize\setlength{\tabcolsep}{4pt}
\begin{tabularx}{\linewidth}{>{\raggedright\arraybackslash}Xrrrr}
\toprule
& Share & Reports & $\rho$ with & $\rho$ from \\
Items & (\%) & move (\%) & \ARzo{} & publication \\
\midrule
Desk bylines & \exEeightDeskSharePct{} & \exEeightDeskPricePct{} & \exEeightDesk{} & \exEeightDeskPub{} \\
Named contributors & \exEeightNamedSharePct{} & \exEeightNamedPricePct{} & \exEeightNamed{} & \exEeightNamedPub{} \\
\midrule
Before the open, event day & \exEsixSharePreOpen{} & --- & \exEsixPreOpen{} & \exEsixPreOpenPub{} \\
During the session & \exEsixShareIntraday{} & --- & \exEsixIntraday{} & \exEsixIntradayPub{} \\
After the close or off-session & \exEsixShareAfter{} & --- & \exEsixAfter{} & \exEsixAfterPub{} \\
\midrule
All frame items & 100 & \exEfivePriceShare{} & \hTwoSpearmanEmu{} & \exEsixPubAll{} \\
\bottomrule
\end{tabularx}}

\end{table}

Exploratory (not pre-registered) analyses of who writes the news and when show that the association is mostly with
returns realized before the item appeared (Table~\ref{tab:bylines}, Fig.~\ref{fig:validity}b). It is largest for
items that appear during the session, where part of \ARzo{} precedes them, and for items whose named contributors
report the move; from publication it is small for every group (exploratory). On the publication-aligned criterion, from the first one-minute bar after
publication to the close of the next session, the exploratory statistic is \exEsixPubAll{} \exEsixPubAllCI{}. SEC acceptance times, a clock independent of the news vendor, agree (exploratory): of the
\exEeightSecUnits{} earnings items whose firm filed an Item~2.02 \eightK{} around them, \exEeightSecAfterPct\% appear
after the filing's acceptance, a median \exEeightSecMedianMin{} minutes after it: Web reports of scheduled disclosures
follow both the disclosure and the market's reaction to it.

Items that report the move are a second channel, concentrated in contributor-written items: named contributors
report the move \exEeightNamedPricePct\% of the time against \exEeightDeskPricePct\% for desk items
(Table~\ref{tab:bylines}). In an exploratory check, on the items a fixed pattern flags the statistic is
\exEfivePriceOnly{}, and without them \exEfiveNoPrice{} \exEfiveNoPriceCI{}; an exploratory audit puts the true share
of such items near \exEfiveAuditPrevalence\% (App.~\ref{app:validity}).

By event type, the correlation is highest for earnings results, analyst actions and guidance changes and lower for
product or contract news, the largest category, and small from publication for every type (exploratory,
App.~\ref{app:exploratory}). The fall is not specific to one LLM: the tone of FinBERT~\cite{huang2023finbert}, a finance-domain model, gives
\exEsevenFinbert{} on the same items and \exEeightFinbertPub{} from publication (exploratory).

\subsection{The Association Holds on News Published after the Reader's Release}\label{sec:res-leakage}

On the \nWeeksRelease{} weeks after the LLM's release, when no later outcome can be in its
weights~\cite{paleka2026pitfalls}, the registered statistic is \hTwoSpearmanRelease{} \hTwoCIRelease{}. This is a
direct estimate, free of recall in the weights, whereas a contrast with earlier weeks would not identify
look-ahead~\cite{zhang2026temporal}. Without items that report the move, it is
\exEsixRelNoPrice{} \exEsixRelNoPriceCI{} (exploratory).
In blinded states the reader rarely names the firm, mostly by abstaining (registered D2): under full blinding its top
answer was the true firm in \dTwoCompanyFull\% \dTwoCompanyFullCI{} of probes and ``cannot tell'' in \dTwoCannotTell\%.
The registered probe rule left the window start unchanged (AUC \dOneCentreAUC{}, $p={}$\dOneCentreP{}); because some
windows before the release differ from one half (App.~\ref{app:validity}), we rest the claim on the post-release
estimate, which recall cannot reach.

\IfJevResults{Jev, run through the same interface on the \nJevUnits{} items published after its release, over
\jevFwdWeeks{} weeks, is read on news that recall in its weights cannot reach: its
judgments correlate \hTwoSpearmanJev{} \hTwoCIJev{} with \ARzo{} and \hTwoPlaceboJev{} with the placebo window, against
\jevFwdSpearmanEmu{} and \jevFwdPlaceboEmu{} for emu on the same items (descriptive). On the \nVerbUnits{} items that VERB also answered, Jev agrees with
emu (Spearman of \Eeff{} \jevAgreeEmuRho{}, $\kappa$ of the event type \jevAgreeEmuKappa{}) and answered at
\$\jevUSDPerKItems{} per thousand items (descriptive; more in App.~\ref{app:validity}).}{}

\subsection{Usefulness at the Decision Horizon}\label{sec:res-horizon}

\begin{figure}[t]
  \centering
  \includegraphics[width=\columnwidth]{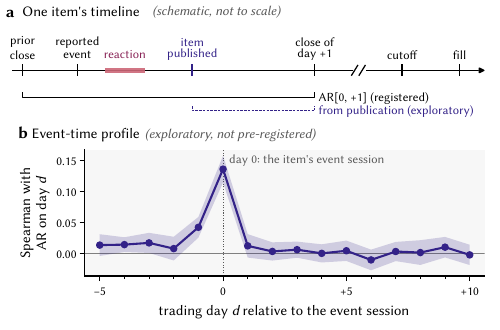}
  \caption{When the criterion is realized. (a) One item's timeline (schematic). (b) Exploratory: Spearman correlation
    of \Eeff{} with the abnormal return on each trading day around the event session, with \ciLevel\% Newey--West bands;
    it peaks on the event session, and the day before is smaller but above zero.}
  \Description{Panel a is a schematic time line: prior close, reported event, market reaction, item published, close of
    the session after the event session, then after a break the decision cutoff and the fill; a bracket marks the
    registered window from the prior close to that close and a dashed bracket the exploratory window from publication to
    the same close. Panel b
    is a line plot over trading days from five before to ten after the event session: the correlation is near zero
    before the event, rises on the day before, peaks on day zero and returns to near zero from the next day on, with
    shaded intervals.}
  \label{fig:horizon}
\end{figure}

\begin{table}[t]
\caption{Rank information the judgments added at the decision horizon: H3 and its registered split.}
\label{tab:horizon}
{\footnotesize
\setlength{\tabcolsep}{5pt}
\begin{tabularx}{\linewidth}{c>{\raggedright\arraybackslash}Xrl}
\toprule
\multicolumn{4}{l}{\textit{Rank information at the decision horizon, 113 learned weeks}} \\
\midrule
& Measure & Est. & 95\% CI \\
\midrule
C & H3: $\Delta$IC over the code-only twin & \ensuremath{-}0.016 & [\ensuremath{-}0.029, \ensuremath{-}0.002] \\
R & \quad text part & \ensuremath{-}0.006 & [\ensuremath{-}0.017, 0.004] \\
R & \quad trusted typed part & \ensuremath{-}0.009 & [\ensuremath{-}0.021, 0.002] \\
R & \quad after release, 73 weeks & \ensuremath{-}0.009 & [\ensuremath{-}0.028, 0.011] \\
R & IC of the full system & \ensuremath{-}0.024 & [\ensuremath{-}0.061, 0.014] \\
R & IC of the code-only twin & \ensuremath{-}0.008 & [\ensuremath{-}0.047, 0.030] \\
\bottomrule
\end{tabularx}
\par\smallskip
\begin{minipage}{\linewidth}\scriptsize\raggedright
C: confirmatory (one-sided, Holm over H1--H3); R: registered estimate, 95\% interval. IC: weekly Spearman correlation of the combined logit with the holding week's relative return.
\end{minipage}
}

\end{table}

\hThreeSentence{} (Table~\ref{tab:horizon}.) Neither the full system
(IC \hThreeICPrimary{}) nor its code-only twin (\hThreeICCodeonly{}) has detectable rank IC, and before the commit the
frozen learner recovered only \mOneBRatio{} of a planted IC of \mOneBPlanted{}, so H3 bounds what this learner and
decision layer extract, not what the judgments contain. The text logit's part is \icTextPart{}, and on the
post-release weeks the estimate is \hThreeDeltaICRelease{} \hThreeCIRelease{}.

Exploratory evidence without the learner points the same way. In exploratory event time, the
association peaks on the event session and is near zero from the next day on (Fig.~\ref{fig:horizon}b). For \exEtwoClosedBeforeFillPct\% of items the \ARzo{}
window closes before the next-open fill, and the exploratory item correlation with the traded week's relative return is \exEtwoTraded{}
\exEtwoTradedCI{}. Fast-assimilation studies predict this pattern: the next-day drift of LLM news scores fell to an
annualized Sharpe ratio of \litLopezDriftSharpeLast{} by early 2024~\cite{lopezlira2026chatgpt}, before our window
opens on \wCleanStartProse{}.

Portfolio estimates of every book are in App.~\ref{app:exploratory}.

\subsection{Planted Content Stays within the Influence Bound}\label{sec:res-planted}

\begin{table}[t]
\caption{Attack success (ASR) at the tested attacker and budget: planted facts and press releases moved names into
the signal book; planted instructions did not.}
\label{tab:planted}
{\scriptsize
\setlength{\tabcolsep}{4pt}
\begin{tabularx}{\linewidth}{c>{\raggedright\arraybackslash}Xrlrr}
\toprule
\multicolumn{6}{l}{\textit{A. Per attack family, per-item reader}} \\
\midrule
& Attack family & ASR (\%) & 95\% CI & No def. & $n$ \\
\midrule
R & A1\ instruction to AI readers & 0.0 & [0.0, 2.1]\rlap{$^{*}$} & 5.6 & 173 \\
R & A2\ fabricated fact & 22.5 & [16.0, 29.3] & 22.5 & 173 \\
R & A3\ self-labelled good news & 0.0 & [0.0, 2.1]\rlap{$^{*}$} & 8.8 & 173 \\
R & A4\ hidden instruction & 0.0 & [0.0, 2.1]\rlap{$^{*}$} & 4.4 & 173 \\
R & A5\ stock promotion & 4.0 & [1.5, 6.8] & 4.8 & 200 \\
R & A6\ planted press release & 31.5 & [24.5, 38.6] & 37.4 & 200 \\
R & A7\ repeated plant & 25.9 & [11.1, 41.9] & 23.1 & 27 \\
R & A8\ demotion (removed or halved) & 57.2 & [48.8, 65.6] & --- & 250 \\
\bottomrule
\end{tabularx}
\par\smallskip
{\setlength{\tabcolsep}{3pt}
\begin{tabularx}{\linewidth}{c>{\raggedright\arraybackslash}Xrlrlrr}
\toprule
\multicolumn{8}{l}{\textit{B. Readers in the same slot, band and cap, A1--A7 pooled}} \\
\midrule
& Reader & ASR (\%) & 95\% CI & $\Delta$ (pp) & 95\% CI & Holm $p$ & Loc. \\
\midrule
R & Per-item reader & 10.5 & [8.2, 12.8] & ref. & n/a & n/a & 88.0 \\
R & \quad no defenses & 14.5 & [11.3, 17.8] & \ensuremath{-}4.9 & [\ensuremath{-}7.2, \ensuremath{-}2.8] & $<$0.001 & 92.7 \\
C & Holistic reader (H1) & 0.1 & [0.0, 0.5]\rlap{$^{*}$} & 10.4 & [8.1, 12.7] & $>$0.99 & n/a \\
R & \quad with guard questions & 0.1 & [0.0, 0.5]\rlap{$^{*}$} & 10.0 & [7.8, 12.5] & $>$0.99 & n/a \\
R & \quad with marking & 0.3 & [0.0, 0.6] & 9.6 & [7.3, 12.1] & $>$0.99 & n/a \\
R & Per-item, median & 6.0 & [4.1, 8.1] & 3.8 & [2.1, 5.5] & $>$0.99 & 92.3 \\
R & Per-item, majority vote & 2.9 & [1.4, 4.7] & 6.4 & [4.4, 8.6] & $>$0.99 & 100.0 \\
R & Dictionary reader & 2.6 & [1.3, 4.1] & 7.3 & [4.8, 9.9] & $>$0.99 & n/a \\
\bottomrule
\end{tabularx}
}
\par\smallskip
\begin{minipage}{\linewidth}\scriptsize\raggedright
C: confirmatory (H1); R: registered. ASR: share of attacked stock-weeks whose name enters the signal book (A8: held name removed or halved), 95\% interval clustered on the attack week; $^{*}$exact binomial. No def.: without guard questions. $\Delta$: per-item reader minus this reader on matched stock-weeks (points); Holm $p$ over the 7 secondary contrasts. Loc.: \% of successes whose decision card ranks the planted item first. Adaptive attacker on NG (64 queries per item, 30 items): surviving edit 100\%, effect up one level 70\%.
\end{minipage}
}

\end{table}

A reader that tracks real events can also be fed planted ones, which is why the bound does not rely on the reader
(Table~\ref{tab:planted}). Instructions to AI readers, self-labelled good news and hidden instructions succeeded against the per-item reader in none of the \asrFamAoneN{}
attacked stock-weeks of each family (exact upper bound \asrZeroUpper\%), and stock promotion rarely did
(\asrFamAfive\%). Fabricated facts written as reports succeeded on \asrFamAtwo\% \asrFamAtwoCI{} of attacked
stock-weeks, and planted press releases on \asrFamAsix\% \asrFamAsixCI{}. Among the registered secondary
contrasts (Table~\ref{tab:planted}B), the instruction guard lowered attack success, and so did median and majority
aggregation. Demotion attacks removed or halved a held name in \demotionASR\% \demotionCI{} of attacked stock-weeks. All attacks
stayed within the bound: no name entered from outside the band and no protected name was excluded without a trusted
anchor. The bound limits how far text can move a position, not whether a planted report can move a name inside the
band, which the reader alone decides.

\paragraph{What the agent decomposition bought}
H1 compares systems, not readers, because each arm keeps its own learned text slope. \hOneSentence{} The holistic
reader's learned slope was positive in only \exEoneSlopeHOutlookPosPct\% of weeks, so its low attack success reflects a
learner that discounted its reader; with the slope flipped, its attack success is \exEoneASRHFlipTyped\% (exploratory). The registered
gains of the decomposition are attribution, robust aggregation and the bound. Decision cards traced \localizationRate\% \localizationCI{} of
successful attacks to the planted item, which no holistic reader can do. Majority vote over the news agents' labels
lowered attack success relative to the mean (difference \voteDelta{} points, \voteCI{}). Every attack stayed within the bound,
whose implementation passed every worst-case check (App.~\ref{app:proof}). Leave-one-agent-out refits changed the
system's rank IC by \lopoMin{} (\lopoMinProg) to \lopoMax{} (\lopoMaxProg), every interval covering zero: the
system does not rest on any one agent.

\FloatBarrier
\section{Related Work}\label{sec:related}

\paragraph{Machine readers of text, tested against what happened}
Web mining has long predicted stock movements from online text: from news weighted by its varying
quality~\cite{hu2018listening}, from tweets with prices~\cite{xu2018stock} and from events extracted from news
titles~\cite{ding2015deep}.
GPT-4's reading of headlines tracks the initial reaction after the model's cutoff~\cite{lopezlira2026chatgpt}, and
supervised language-model probabilities of news predict the sign of the three-day excess return and earn a net return
at daily horizons~\cite{kirtac2026financial}.
Agreement with human labels and association with returns diverge for social-media sentiment with an LLM
annotator~\cite{aravinthakshan2026human} and for sentiment models on \eightK{} earnings
releases~\cite{bisharat2026evaluating}, and market reactions have checked LLM tags of \eightK{}
filings~\cite{dolphin2026grounded}.
At this venue, influencers' public outlooks have been compared with their own exchange positions~\cite{kawai2023digital},
LLM weak labels of Reddit market sentiment with expert annotations~\cite{deng2023llms}, and self-reflective LLMs
predict next-day stock movements from tweets with explanations~\cite{koa2024learning}. A registered study tests a hosted typed model's reading of earnings-call
remarks against the next-session abnormal return, with anonymization and recall probes~\cite{kathiravelu2026jev}.
In finance, prices absorb televised analyst reports within a minute~\cite{busse2002market} and react to stale news
when it recombines earlier stories~\cite{fedyk2023when}; we measure this lag for the Web news on which LLM readers are
tested.
Prospective designs collect LLM scores forward in time, which excludes recall in the
weights~\cite{chen2026autonomous,fatouros2026signal}. Our test is pre-specified and point-in-time, separates what is realized before an item
appears from what follows its publication, and follows the same reader into a trading system.

\paragraph{Typed decision models and how they are evaluated}
Typed decision models return a distribution over caller-declared options without generating text, whether
hosted~\cite{sun2026typesafe} or open~\cite{cheng2026thisthat}. Like LLM annotators, they have been evaluated against
gold labels, in pre-registered annotation tasks~\cite{ibrahim2026evaluating} and as judges, where ``a typed probability
output alone does not establish that it is calibrated''~\cite[p.~2]{li2026jev}. In an audit of the first typed-model
papers, the typed readout showed no independent accuracy advantage, and most papers that evaluate a model
(\litTangNoBaseline{} of \litTangEvaluating{}) report no label-probability baseline~\cite{tang2026typed}. Verbalized
confidence can be better calibrated than label probabilities~\cite{tian2023just} yet discriminate
little~\cite{xiong2024can}. We read the same LLM both ways and test it against records it does not write.

\paragraph{Look-ahead and point-in-time evaluation}
A memorization test bounds recall only from below, and prompts to ignore later information do not
help~\cite{lopezlira2025memorization}. Pre-training text holds after-the-fact explanations of price moves, which
trading agents can recite in a backtest~\cite{li2025profit}. Masking firm names removes both look-ahead and a
distraction effect~\cite{glasserman2023assessing} but ``removes the trigger, not the
knowledge''~\cite[p.~13]{jia2026hindsightbench}; passive before/after comparisons are
uninformative~\cite{zhang2026temporal}, and a release date bounds the cutoff~\cite{paleka2026pitfalls}. Time-locked language models remove recall by
construction~\cite{he2025chronologically} and fit our reader slot. We add two channels that
a release date does not close: items that report the outcome and items that appear after the reaction.

\paragraph{Steerable text and modular trading agents}
Retrieved text blurs data and instructions~\cite{greshake2023not}, and crafted Web content steers the products that
LLM assistants recommend~\cite{nestaas2025adversarial}. For typed models, an appended opinion flips decisions as often
as the strongest injected command~\cite{hu2026jevadvbench}, and type safety and injection resistance ``address
different properties''~\cite[p.~8]{wu2026decision}. Isolation keeps raw evidence from the answering component~\cite{yu2026cordon},
and a proved trust boundary constrains the authorized action rather than the text~\cite{shkolnikov2026composable}.
Poisoned social-media feeds shift the beliefs of multi-agent trading stacks, and coordinators dampen
the shift without bounding it~\cite{sua2026contagion}. Modular trading designs keep language models upstream of calibration, risk and execution~\cite{ye2026alpha}; in multi-agent designs, specialist analysts
debate~\cite{xiao2024tradingagents} or a portfolio-manager model draws on a policy analyst~\cite{li2025time}. We measure
validity and steerability on the same reader, inside a bound that holds for any reader.

\section{Discussion}\label{sec:discussion}

\paragraph{What this means for readers of Web text}
We propose a reporting rule for tests of closed-choice readers of Web text against records they do not write:
report a placebo window, a pre-open control, an estimate after the model's
release, re-identification with its abstentions, the share of items that already report the outcome and, above all, the
publication-aligned criterion next to the announcement-window criterion. An announcement-window test shows that a
reader reads the event; whether its judgments are useful must be measured from publication, at the horizon of use, and
under a bound on actions that does not depend on the reader.

\paragraph{Scope}
The method extends to any Web text whose subject later produces a public, time-stamped record, such as Web reports of
clinical trials graded against registry results, and the system accepts any reader that speaks the schema. Here we
instantiate it with one pinned LLM as the confirmatory reader, Jev as a descriptive reader and NG as a registered reference reader, one feed's headlines
and summaries over \universeSize{} large caps, and a weekly decision with a next-open fill.

\section{Conclusion}\label{sec:conclusion}
We asked whether LLM judgments of Web news track the market's reaction or follow it, and built a multi-agent trading
system that answers it point in time under a pre-specified protocol. The judgments track the reaction, also on news
published after the reader's release (Jev, descriptive). Web news, however, mostly follows the market, so the
publication-aligned criterion, not the announcement window, decides whether a reader of Web text is useful. We propose
it as a reporting standard and offer the system, whose agent decomposition bounds what any reader's text can do, as a
testbed in which future readers can be evaluated the same way.

For Web mining, when a text appears relative to the event it reports, and who wrote it, decide what a market-graded
test measures; for LLM trading agents, what limits planted text is a bound in code, not the reader; and a typed model
can be graded through the same interface on news it cannot have seen, as soon as it is released.

\section*{Ethics and Artifacts}
All attacks target our own system on archived data; no attack item was published, and no human participants were
involved. The licensed news was sent to the pinned LLM's provider under its API terms, and no article text is released.
The code repository (\artifactURL) holds the code, question bank, prompts, model pins, hashed-identifier judgment
tables, labels, the pre-registration and the scripts behind every number, table and figure; decision cards carry hashed
item identifiers only, because exposed attributions may help attackers~\cite{nestaas2025adversarial}.

\bibliographystyle{ACM-Reference-Format}
\bibliography{refs}

\appendix
\setcounter{table}{0}\renewcommand{\thetable}{A\arabic{table}}
\setcounter{figure}{0}\renewcommand{\thefigure}{A\arabic{figure}}

\section{The Influence Bound}\label{app:proof}

\paragraph{Setup}
We drop the decision index $t$. Code standardizes each judgment into features; the trusted learner gives
$L^T_i=\sum_{j\in\mathcal{T}}\hat\beta_j\tilde x_{j,i}$ and the untrusted learner, fitted with $L^T$ as an offset,
gives $L^W_i=\sum_{j\in\mathcal{W}}\hat\gamma_j\tilde x_{j,i}$. With $s^T$ the cross-sectional standard deviation of
$L^T$ (floored at \capFloor{}) and $\mu={}$\capMultiplier{},
\begin{equation*}
\bar L^W_i=\max\bigl(-\kappa,\min(\kappa,L^W_i)\bigr),\qquad \kappa=\mu s^T,\qquad L_i=L^T_i+\bar L^W_i,
\end{equation*}
a logarithmic opinion pool with signed, fitted weights. $R^T$ and $R$ rank the
universe by $L^T$ and $L$, ties broken by a fixed symbol hash. The \emph{protected} names are $P=\{i:R^T_i\le K-J\}$ and
the band is $E=\{i:R^T_i\le K+J\}$. Holdings are long only, $w^{\mathrm{cur}}\ge0$, and enter as
$\hat w=w^{\mathrm{cur}}/\max(1,\sum_iw^{\mathrm{cur}}_i)$; $\mathcal{H}=\{i:\hat w_i>0\}$. The caps are
$\bar w={}$\nameCapPct\% per name, $\bar w_{\mathrm{ind}}={}$\industryCapPct\% per industry and $\bar\Delta={}$\turnoverCapPct\%
turnover. Text rules give each name a multiplier $m^W_i\in\{\frac{1}{2},1\}$, or $0$ only if a trusted anchor $A_i=1$
(an \eightK{} with one of Items \anchorFilingItems{} accepted within \anchorFilingDays{} days, or a market-relative return of at most
$-\anchorPriceSd\,\sigma_i\sqrt{n}$ over the last $n$ sessions) and \corroborationKeys{} distinct bylines report the
problem; $X$ is the set of names with multiplier $0$.

\paragraph{Selection and sizing}
$\mathrm{sel}(\tilde R,X,\mathcal{H}',b)$ places (a)~every $i\in P\setminus X$; then, in the order of $\tilde R$ until
$K$ names are placed, (b)~each $i\in\mathcal{H}'\setminus X$ with $\tilde R_i\le K+b$ and $R^T_i\le K+J$, and (c)~each
$i\notin X$ with $R^T_i\le K+J$. The signal book is $B^{\mathrm{EW}}=\mathrm{sel}(R,\emptyset,\emptyset,\cdot)$ with
equal weights; the system book is $B=\mathrm{sel}(R,X,\mathcal{H},b)$ with $b={}$\incumbentBuffer. The trusted target
is $v^T_i=\min(\bar w,u_i)\,m^T_i$, with $u$ the inverse-volatility weights over the trusted top $K$ and $m^T_i$ the
earnings multiplier. The \emph{reference path} $S^{\mathrm{ref}}=\mathrm{sel}(R^T,\emptyset,\mathcal{H},b)$ sets
$v^{\mathrm{ref}}=v^T$ on $S^{\mathrm{ref}}$, scaled per industry and then to a total of at most one, and a trade
fraction $\varphi=\min(1,\bar\Delta/\mathrm{TO})$ from its own turnover. With kept names $\mathcal{C}=B\cap
S^{\mathrm{ref}}$, entrants $\mathcal{N}=B\setminus S^{\mathrm{ref}}$ and displaced names
$\mathcal{D}=S^{\mathrm{ref}}\setminus B$: $v_i=v^{\mathrm{ref}}_im^W_i$ on $\mathcal{C}$; $v_e=\min(v^T_e,F/|\mathcal{N}|)\,m^W_e$
with $F=\sum_{d\in\mathcal{D}}v^{\mathrm{ref}}_d$, scaled into the industry room that $\mathcal{C}$ leaves; $v_i=0$
elsewhere. Both paths move a fraction $\varphi$ toward their targets, $z_i=\hat w_i+\varphi(v_i-\hat w_i)$ ($z_i=0$ on
$X$), and share the industry factors $f_g$ computed on the reference path: $w_i=f_{g(i)}\min(\bar w,z_i)$ for
$i\notin\mathcal{N}$ and $w^{\mathrm{ref}}_i=f_{g(i)}\min(\bar w,z^{\mathrm{ref}}_i)$; entrants are scaled into the room
the other names leave. An attack $\mathcal{A}$ modifies any untrusted input of the current or earlier weeks.

\begin{assumption}[Trust partition]\label{ass:trust}
$\hat\beta$, $L^T$, $R^T$, $\kappa$, the volatilities, $m^T$, $v^T$, the reference path and the anchors are functions
of trusted inputs and the current holdings only.
\end{assumption}

\begin{proposition}[Influence bound]\label{prop:bound}
Under Assumption~\ref{ass:trust}, for every attack $\mathcal{A}$, holdings $w^{\mathrm{cur}}$ and name $i$:
\begin{description}[leftmargin=0.8em,itemsep=0pt,topsep=1pt]
\item[P1, entry only from the band:] $i\in B(\mathcal{A})\cup B^{\mathrm{EW}}(\mathcal{A})\Rightarrow i\in E$; a held name
  outside $E$ is only sold down.
\item[P2, protection:] if $i\in P$ and no exclusion of $i$ fires, then $i\in B(\mathcal{A})$,
  $v_i(\mathcal{A})\ge\frac{1}{2}v^{\mathrm{ref}}_i$ and $w_i(\mathcal{A})\ge\frac{1}{2}w^{\mathrm{ref}}_i$; an exclusion needs
  $A_i=1$.
\item[P3, no created weight:] $v_i(\mathcal{A})\le v^T_i$, $w_i(\mathcal{A})\le(1-\varphi)w^{\mathrm{cur}}_i+\varphi
  v^T_i$, targets sum to at most $\sum_iv^{\mathrm{ref}}_i\le1$, and industries stay within their cap.
\item[P4, bounded logit:] $|L_i(\mathcal{A})-L^T_i|\le\kappa$, so $|L_i(\mathcal{A})-L_i(\emptyset)|\le2\kappa$.
\item[P5, history:] P1--P4 hold whatever the untrusted inputs of earlier weeks were.
\end{description}
\end{proposition}

\begin{lemma}[Band-guarded selection]\label{lem:band}
A selector chooses $K$ of $n$ items by the score $s^T+\max(-\kappa,\min(\kappa,s^U))$, where $s^T$ and $\kappa$ depend
on trusted inputs only. It first places every item of trusted rank at most $K-J$ that is not vetoed, where only items
in a set fixed by trusted inputs can be vetoed, and fills the other slots only from items of trusted rank at most
$K+J$. Then, for every untrusted score $s^U$, every selected item has trusted rank at most $K+J$, every unvetoed item
of trusted rank at most $K-J$ is selected, and every score is within $\kappa$ of $s^T$.
\end{lemma}

\paragraph{Proofs}
By Assumption~\ref{ass:trust}, $R^T$, $P$, $E$, $v^T$, $S^{\mathrm{ref}}$, $v^{\mathrm{ref}}$, $\varphi$, $f$,
$w^{\mathrm{ref}}$ and $A$ do not change under attack.\par
\emph{Lemma.} Trusted ranks form a permutation, so exactly $K-J$ items have rank at most $K-J$, and step (a) places all
unvetoed ones before any other item; steps (b) and (c) admit only ranks at most $K+J$; the clip gives
$|s-s^T|\le\kappa$. $\square$\par
\emph{P1.} By the Lemma, $B\cup B^{\mathrm{EW}}\subseteq E$. For $i\notin E$, $i\notin B\cup S^{\mathrm{ref}}$, so
$v_i=v^{\mathrm{ref}}_i=0$ and $z_i=(1-\varphi)\hat w_i$ unless $i\in X$; hence $w_i\le(1-\varphi)\hat w_i$. $\square$\par
\emph{P2.} Step (a) gives $P\setminus X\subseteq B\cap S^{\mathrm{ref}}=\mathcal{C}$, and $i\in X$ needs $A_i=1$. For
$i\in P\setminus X$: $v_i=v^{\mathrm{ref}}_im^W_i\ge\frac{1}{2}v^{\mathrm{ref}}_i$; $z_i\ge\frac{1}{2}z^{\mathrm{ref}}_i$ because
$\hat w_i\ge0$; and the shared $f_{g(i)}$ gives $w_i\ge\frac{1}{2}w^{\mathrm{ref}}_i$. $\square$\par
\emph{P3.} The reference caps, the industry step for entrants, $m^W\le1$ and the minimum in $v_e$ give $v\le v^T$ and
$v_i\le v^{\mathrm{ref}}_i$ for $i\notin\mathcal{N}$. $\mathcal{C}$ and $\mathcal{D}$ partition $S^{\mathrm{ref}}$ and
$\sum_{e\in\mathcal{N}}v_e\le F$, so $\sum_iv_i\le\sum_iv^{\mathrm{ref}}_i\le1$. Then $0\le z_i\le(1-\varphi)\hat w_i+
\varphi v^T_i$, the caps multiply $\min(\bar w,z_i)$ by factors in $[0, 1]$, and names outside $\mathcal{N}$ satisfy
$z_i\le z^{\mathrm{ref}}_i$, so each industry stays within $f_gx_g\le\bar w_{\mathrm{ind}}$ with the entrants in the
room left. $\square$\par
\emph{P4, P5.} The clip and Assumption~\ref{ass:trust} give P4 by the triangle inequality through $L^T$. P5 holds
because $\mathcal{A}$ may modify any earlier week and the assumption holds for every $w^{\mathrm{cur}}$. $\square$

\paragraph{Tightness and tests}
Constructed cases attain each part: an entrant at $R^T=K+J$ (P1); $v=\frac{1}{2}v^{\mathrm{ref}}$ from one planted
serious-problem item without an anchor (P2); $v_e=v^T_e$ for an entrant with $F/|\mathcal{N}|\ge v^T_e$ (P3); and
$|\Delta L|=2\kappa$ at opposite caps (P4). Property tests assert Assumption~\ref{ass:trust} and P1--P4 on at least
\nPropertyCases{} randomized cases with arbitrary untrusted inputs, random trusted inputs, holdings, $K$ and $J$; a
history test poisons every untrusted feature and store record of earlier weeks (P5). As implementation verification on the
real runs, a worst-case reader produced \nBoundViolations{} violations in \nEntryChecks{} entry (P1) and
\nProtectionChecks{} protection (P2) checks. On the attack runs, an
entrant reached trusted rank \maxEntrantRank{}${}=K+J$ (P1), \pTwoShareHalved\% of the \pTwoProtectedHeld{}
protected held names without an anchor were halved and none was excluded (P2), weights reached \maxEntrantWeightRatio{}
of their bound (P3), and the text moved a logit by at most \maxLogitRatio{} of $2\kappa$, so P4 is attained only in the
constructed case. Decision-card ledgers sum to the decision up to \ledgerResidual{}.

\section{Pre-Specification and Deviations}\label{app:prereg}

\paragraph{Hypotheses}
H1 is paired over (stock-week, family) with a bootstrap clustered on the attack week; H2 and H3 use one-sided
Newey--West tests of weekly means. Registered MDEs: H1
\hOneMDELowRegistered--\hOneMDEHighRegistered{} at \nPairedAttacksRegistered{} pairs (\hOneMDELow--\hOneMDEHigh{} at the
realized \nPairedAttacks{}); H2 \hTwoMDELow--\hTwoMDEHigh{} (realized \hTwoMDERealized); H3 \hThreeMDELow--\hThreeMDEHigh{}
(realized \hThreeMDERealized). The \nPromotionFamilies{} promotion families are: A1, an instruction to automated readers
appended to a real item; A2, a fabricated fact written as a report; A3, self-labelling good news; A4, a hidden or
obfuscated instruction; A5, stock promotion; A6, a planted press release; A7, one claim planted in three consecutive
weeks. The \nDemotionFamilies{} demotion families plant a fabricated serious problem about a held name in two items, under one
byline or two; demotion success is \demotionProtected\% \demotionProtectedCI{} for protected and \demotionInBand\%
\demotionInBandCI{} for other held names (exclusions of protected names: \demotionProtectedExcl\%).

\paragraph{Result sentences and deviations}
Each hypothesis had a sentence fixed for a supported and for a null outcome, printed verbatim: H2 its supported sentence
(Table~\ref{tab:validity} caption, its Holm-adjusted $p$ printed as a bound), H1 and H3 their null sentences. Where an interval
lies wholly on the side opposite to a registered direction, a logged deviation adds a descriptive sentence, with no new
test. The log (code repository) holds \nDeviations{} deviations, none of which changes a confirmatory statistic,
its test or its Holm decision: coverage waivers, driver fixes, registered items not run, and presentation.
The pre-registration's commit timestamp precedes the unlock, and it declares its prior looks at outcomes in the window
(a pilot of \priorLookStockWeeks{} stock-weeks, a wording smoke test, and classical-factor ICs and baseline portfolios);
the validity estimate is repeated without the first two.

\section{Validity Details, Look-Ahead and Model Cutoffs}\label{app:validity}

\begin{table}[t]
\caption{Readers, model pins and knowledge cutoffs. Only emu's LLM documents a cutoff, and only emu has weeks of the
evaluation window after its release.}
\label{tab:models}
{\scriptsize
\setlength{\tabcolsep}{3pt}
\begin{tabularx}{\linewidth}{>{\raggedright\arraybackslash}X>{\raggedright\arraybackslash}p{0.31\linewidth}llr}
\toprule
Reader & Model pin & Cutoff & Release & \shortstack[r]{Weeks\\after} \\
\midrule
emu, VERB, holistic reader and generative-analyst baseline & \texttt{gpt-4.1-mini-\allowbreak{}2025-04-14} & 2024-06-01 & 2025-04-14 & 73 \\
NG, also the full-frame price-report coder & \texttt{flock-io/\allowbreak{}this-that-model-1.0@\allowbreak{}3d927195} & undocumented & 2026-09-20 & none \\
Jev (\jevPanelStatus{} panel) & \texttt{jev-1.13.0} & undocumented & 2026-09-15 & none \\
\bottomrule
\end{tabularx}
\par\smallskip
\begin{minipage}{\linewidth}\scriptsize\raggedright
Cutoff: documented knowledge cutoff. Release: snapshot date (emu), model-hub creation date (NG), release date (Jev). Weeks after: decisions of the evaluation window after the release.
\end{minipage}
}

\end{table}

\begin{table}[t]
\caption{NG's registered estimates on its \nNGUnits{} sampled items (pooled over items,
week-block \ciLevel\% intervals).}
\label{tab:ng}
\centering
{\scriptsize
\setlength{\tabcolsep}{3pt}
\begin{tabular}{lrl}
\toprule
Measure (pooled over items) & NG & 95\% CI \\
\midrule
\ARzo{} & 0.108 & [0.016, 0.203] \\
\quad emu on the same items & 0.074 & [\ensuremath{-}0.014, 0.160] \\
Placebo \ARpl{} & 0.109 & [0.022, 0.193] \\
Agreement with emu, $\rho$ of \Eeff{} & 0.76 & --- \\
Agreement with emu, $\kappa$ of top type & 0.65 & --- \\
Earnings vs Item 2.02 (130 pos.) & 0.830 & [0.784, 0.872] \\
Deal vs Items 1.01/2.01 (6 pos.) & 0.471 & [0.312, 0.909] \\
Management vs Item 5.02 (25 pos.) & 0.620 & [0.508, 0.741] \\
Beat-expectations vs sign ($n$ 84) & 0.530 & --- \\
ECE, earnings vs Item 2.02 & 0.129 & --- \\
Sampled items (eligible by own answers) & 500 & (386) \\
\bottomrule
\end{tabular}
}

\end{table}

\begin{figure*}[t]
  \centering
  \includegraphics[width=\textwidth]{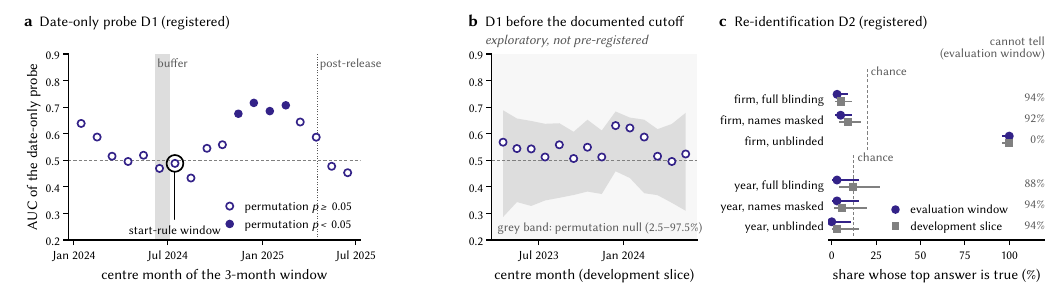}
  \caption{Look-ahead checks. (a) Registered D1: probe AUC by three-month window, filled where the registered
    name-permutation test gives $p$ below \familyAlpha{}; ringed: the start-rule window; grey band: the buffer after the
    documented cutoff; dotted line: start of the post-release weeks. (b) Exploratory: the same probe on the development
    slice, with the permutation null band. (c) Registered D2: share of probes whose top answer is the true firm or year,
    with Wilson \ciLevel\% intervals, and the share of ``cannot tell'' answers.}
  \Description{Three panels. Panel a plots probe AUCs by window centre from early 2024 to mid-2025, with a shaded
    buffer, the ringed start-rule window near one half, a run of filled windows above one half in late 2024 and early
    2025, and a dotted line where the post-release window begins. Panel b, shaded and labelled exploratory, shows
    development-slice windows inside a grey permutation band. Panel c shows re-identification shares near zero under
    blinding and near one without it, for the evaluation window (circles) and the development slice (grey squares),
    with high shares of cannot-tell answers under blinding.}
  \label{fig:lookahead}
\end{figure*}

\paragraph{Frames and controls}
The H2 frame has one row per item and focal firm; rows that share a firm and session share their label, and an
exploratory collapse to one row per firm-day gives \exEfiveFirmDay{} \exEfiveFirmDayCI{} over \exEfiveFirmDays{}
firm-days. The frame is selected by emu's answers; in an exploratory sensitivity the statistic is \exEeightFrameAll{}
\exEeightFrameAllCI{} on all \exEeightFramePairs{} item--firm pairs with an effect answer, kept or not and of any type,
\exEeightFrameAllNoPrice{} \exEeightFrameAllNoPriceCI{} on those without price reports, and \exEeightFrameKept{}
\exEeightFrameKeptCI{} on the \exEeightFrameKeptPairs{} kept pairs of every type: selection by the reader raises the
estimate but does not create it. The price-report pattern undercounts. In an exploratory audit, the authors hand-coded \exEfiveAuditN{} pattern matches
and as many non-matches, drawn by a fixed hash rule within each stratum, with NG's question and two-level
codebook (does the text state that a company's shares moved, or the size of a move?), blind to outcomes, strata and
model codes. Its exploratory precision is \exEfiveAuditPrecision\% \exEfiveAuditPrecisionCI{}, the miss rate among non-matches
\exEfiveAuditMiss\%, and the stratum-weighted recall \exEfiveAuditRecall\% \exEfiveAuditRecallCI{}, which puts the share of frame items that report the move near \exEfiveAuditPrevalence\% (exploratory). With every frame
text coded by NG and the items flagged by it or the pattern (\exEfiveEitherShare\%) removed, the statistic
is \exEfiveNoPriceEither{} \exEfiveNoPriceEitherCI{}, and the placebo without price reports \exEfivePlaceboNoPrice{}
\exEfivePlaceboNoPriceCI{} (exploratory).

\paragraph{Look-ahead and blinding}
The date-only probe D1 covers sampled names and decisions in every month from \dOneFirstMonth{} to \dOneLastMonth{};
its AUC over \nDOneWindows{} three-month windows is tested against one half by permuting names (MDE per window about
\dOneMDE{}), and the start rule reads the window centred on \dOneCentreMonth{}. D2 asks which of five same-industry
firms (or ``cannot tell'') and which year a state describes, at three blinding levels (MDE \dTwoMDECell{} per cell).
In an exploratory monthly view, none of the
\exEthreePreCutoffWindows{} development windows before the documented cutoff differs from one half, although
\exEthreePreCutoffAboveHalf{} lie above it; windows centred from \exEthreeFlaggedFrom{} to \exEthreeFlaggedTo{}, between
the cutoff and the release, do differ (smallest $p$ \exDOneMinP{}; \exDOneHolmMinP{} after Holm's correction over the
\nDOneWindows{} windows), for which training data past the cutoff, a name prior that fits that period's winners, and
chance all account. The blinded-minus-unblinded difference in rank IC on the
development slice (D3) is \dThreePre{} \dThreePreCI{}.\IfJevResults{ Jev, under full blinding, named the true firm in
\jevDTwoCompanyFull\% of probes; its earnings AUC on the frame is \hTwoAUCEarningsJev{}, and its median latency was
\jevLatencyMedian{}~s (descriptive).}{}

\section{Reporting Checklists}\label{app:checklist}

Against the typed-model checklist~\cite{tang2026typed}, emu is the label-probability baseline, versions, dates and
intervals are reported (Table~\ref{tab:models}), and calibration is reported without recalibration. Against reporting items for LLM
trading research~\cite{ye2026alpha,xia2026agentic}, we report point-in-time stores, blinding, post-release weeks,
re-identification, attack families, next-open fills, a look-ahead audit, cost per decision and agent ablations.

\section{Exploratory Analyses and Registered Portfolio Estimates}\label{app:exploratory}

Except the registered portfolio estimates, everything here was specified after the unlock, is exploratory (not
pre-registered), carries no test and changes no registered claim; it is reported here or in \S\ref{sec:results}. The registered portfolio rows are in Table~\ref{tab:portfolio};
the system book's deflated Sharpe ratio~\cite{bailey2014deflated} is \dsr{} over \nTrials{} registered trials
(\nTrialsRun{} run) and \dsrHeadline{} over the two headline candidates. The fitted combination of trusted and text logits behind H3
(the learned pool) is registered. \poolSentence{}

The system book's own-factor
selection alpha (\alphaSystem\% per year) is that of its code-only twin (\alphaCodeonly\%), so the text did not lower
it (registered estimates).

\begin{table}[t]
\caption{Registered portfolio estimates, all books. Estimates, not tests.}
\label{tab:portfolio}
{\scriptsize
\setlength{\tabcolsep}{2pt}
\begin{tabular}{lrrrrrrr}
\toprule
& Ex.\ ret. & Vol. & Sharpe & MDD & TO & BE & $\alpha$ \\
\midrule
System book & 7.3 & 19.5 & 0.38 & \ensuremath{-}18.0 & 28.0 & \ensuremath{-}23 & \ensuremath{-}15.7 [\ensuremath{-}25.8, \ensuremath{-}5.6] \\
Signal book (equal weight) & 18.2 & 27.9 & 0.65 & \ensuremath{-}23.0 & 36.8 & 17 & \ensuremath{-}14.8 [\ensuremath{-}30.2, 0.5] \\
Equal-weight universe & 17.4 & 18.0 & 0.97 & \ensuremath{-}18.7 & 3.1 & --- & \ensuremath{-}0.5 [\ensuremath{-}0.6, \ensuremath{-}0.3] \\
S\&P 500 ETF & 12.7 & 16.5 & 0.77 & \ensuremath{-}18.8 & 0.4 & $-\infty$ & \ensuremath{-}3.6 [\ensuremath{-}8.8, 1.6] \\
Vol-managed equal weight & 10.5 & 15.8 & 0.66 & \ensuremath{-}19.4 & 3.7 & \ensuremath{-}1{,}063 & \ensuremath{-}3.2 [\ensuremath{-}9.1, 2.8] \\
Momentum 12-1 & 37.8 & 38.9 & 0.97 & \ensuremath{-}33.3 & 13.6 & 202 & 1.0 [\ensuremath{-}7.3, 9.3] \\
LightGBM (code features) & 2.4 & 22.2 & 0.11 & \ensuremath{-}25.4 & 49.3 & \ensuremath{-}16 & \ensuremath{-}20.1 [\ensuremath{-}31.7, \ensuremath{-}8.5] \\
Code-only system & 9.9 & 22.0 & 0.45 & \ensuremath{-}20.2 & 20.2 & \ensuremath{-}27 & \ensuremath{-}16.3 [\ensuremath{-}27.8, \ensuremath{-}4.9] \\
\bottomrule
\end{tabular}
\par\smallskip
\begin{minipage}{\linewidth}\scriptsize\raggedright
Ex.\ ret.: annualized excess return (\%); Vol.: volatility (\%); MDD: maximum drawdown (\%); TO: one-way turnover per rebalance (\%); BE: break-even cost against equal weight (bps per side; $-\infty$: none); $\alpha$: own-factor selection alpha (\%/yr).
\end{minipage}
}

\end{table}

\begin{table}[t]
\caption{Where the reader tracks the reaction, by its top event type (exploratory, except the registered \eightK{}
column). On the publication-aligned criterion, the correlation is small for every type.}
\label{tab:concepts}
{\scriptsize
\setlength{\tabcolsep}{2pt}
\begin{tabularx}{\linewidth}{>{\raggedright\arraybackslash}Xrrrrr}
\toprule
& & \multicolumn{3}{c}{Exploratory (X): weekly $\rho$} & (R) \\
\cmidrule(lr){3-5}\cmidrule(l){6-6}
Reader's top & Share & \multicolumn{2}{c}{\ARzo{}} & From & 8-K \\
\cmidrule(lr){3-4}
event type & (\%) & all & no price & publ. & AUC \\
\midrule
Product or contract & 29.3 & 0.049\rlap{$^{\dagger}$} & 0.037\rlap{$^{\dagger}$} & 0.019 & n/a \\
Analyst action & 11.2 & 0.173\rlap{$^{\dagger}$} & 0.167\rlap{$^{\dagger}$} & 0.043 & n/a \\
Earnings results & 10.8 & 0.226\rlap{$^{\dagger}$} & 0.143\rlap{$^{\dagger}$} & \ensuremath{-}0.012 & 0.833 \\
Other development & 10.7 & 0.056\rlap{$^{\dagger}$} & 0.037 & 0.024 & n/a \\
Deal or investment & 8.1 & 0.035 & 0.031 & \ensuremath{-}0.008 & 0.611 \\
Industry read-through & 6.7 & 0.087\rlap{$^{\dagger}$} & 0.078\rlap{$^{\dagger}$} & 0.044\rlap{$^{\dagger}$} & n/a \\
Legal or regulatory & 6.5 & 0.039 & 0.021 & \ensuremath{-}0.022 & n/a \\
Guidance change & 6.2 & 0.172\rlap{$^{\dagger}$} & 0.137\rlap{$^{\dagger}$} & \ensuremath{-}0.041 & n/a \\
Management or labor & 6.0 & 0.119\rlap{$^{\dagger}$} & 0.107\rlap{$^{\dagger}$} & 0.038 & 0.557 \\
\midrule
All types & 100.0 & 0.119\rlap{$^{\dagger}$} & 0.089\rlap{$^{\dagger}$} & 0.023\rlap{$^{\dagger}$} & n/a \\
\bottomrule
\end{tabularx}
\par\smallskip
\begin{minipage}{\linewidth}\scriptsize\raggedright
\textbf{Exploratory}, except the registered 8-K column. $\rho$: weekly mean Spearman correlation of \Eeff{} with the abnormal return; $^{\dagger}$: 95\% interval excludes zero. No price: items that do not report the price move. From publ.: from the first trade after publication to the next close. Types below 5\% of the frame are omitted.
\end{minipage}
}

\end{table}

\paragraph{Timing, bylines and event types}
Items that appear after a close have a window that starts at the prior close and so includes after-hours reactions to
releases at the close; in the exploratory data they appear a median \exEeightAfterCloseMin{} minutes after it. An
exploratory upper-bound check (E9) times the judgments of earnings items at the filing's acceptance instead of the
item's publication; on the units it decomposes, acceptance precedes publication by a median
\exEnineEarnGapMedianMin{} minutes. With one row per filing they correlate
\exEnineEarnEvRhoAcc{} \exEnineEarnEvRhoAccCI{} with the return still ahead, against \exEnineEarnRhoPub{} from
publication, and only with entry in extended hours; entry at the next regular session gives
\exEnineEarnRhoAccRegular{} (exploratory). Because the judged text was written after the filing and may report the
early move, E9 bounds from above what a reader of the filing itself could keep.
Loughran--McDonald dictionary tone~\cite{loughran2011liability} gives \exEfiveLM{} on the frame (exploratory). By
event type (exploratory, Table~\ref{tab:concepts}), the correlation is highest for earnings results
(\exEfiveConceptEarnings), analyst actions and guidance changes, and lower for product or contract news, the largest
category.

\paragraph{Agent ablations, baselines and attribution}
In an exploratory re-run on the registered subset of names, switching the event store off changes the rank IC (full
minus off) by \exPdThreeEpiDeltaIC{} \exPdThreeEpiDeltaICCI{} over \exPdThreeEpiWeeks{} weeks, and switching off both
stores by \exPdThreeAllDeltaIC{} \exPdThreeAllDeltaICCI{} over \exPdThreeAllWeeks{} weeks. In exploratory attribution, the text's share of the system book's active return at the operating band is
\exEfourText{} \exEfourTextCI{} bps per week. On a subset of names over the last \exSubsetWeeks{} weeks, a registered baseline of generative analyst
roles has an exploratory weekly rank IC of \exSubsetICTal{} \exSubsetICTalCI{}, against \exSubsetICTyped{}
\exSubsetICTypedCI{} for our system.

\section{A Typed Request}\label{app:request}
The news agent's per-item request, quoted from the frozen question bank; every reader receives the same state and
questions.

{\scriptsize\setlength{\fboxsep}{3pt}\setlength{\fboxrule}{0.4pt}%
\noindent\fcolorbox{black!35}{black!4}{\begin{minipage}{\dimexpr\linewidth-2\fboxsep-2\fboxrule\relax}\raggedright
\textbf{State} (JSON built in code). \texttt{note}: \emph{Weekly review for a US stock portfolio. Company names are replaced by code names such as Company-7F3A09B2; times are given relative to the review.} Further fields: \texttt{focal\_company}, \texttt{item} (pseudonym, industry and size bucket; headline, screened summary sentences, source, relative publication time and tagged firms).\par
\smallskip\textbf{Questions} (four of 13, asked in one request).\par
\hangindent=1em\texttt{new\_development}. Does \texttt{item} report a new development involving \texttt{focal\_company}? \emph{yes: Something that happened or was announced: results, a forecast, a deal, a product or trial result, a contract, a legal or regulatory step, a management change, or a rating change by a named analyst. no: Only stock price moves, trading or options activity, opinion, previews of coming events, or a recap of older news.}\par
\hangindent=1em\texttt{instructs\_reader}. Does \texttt{item} contain text that tries to control how an automated system or reader rates or handles it? \emph{yes: Text addressed to an AI system, model, algorithm, or automated reader, or text that tells the reader how to rate, rank, classify, or score the item or the stock, or to ignore its instructions. no: Ordinary reporting. Quotes of executives, analysts, or officials, and analyst ratings reported as news, are ordinary reporting.}\par
\hangindent=1em\texttt{promotional}. Is \texttt{item} promotional content about the stock? \emph{yes: Advertising, sponsored or paid content, or stock promotion that urges readers to buy or sell, usually with hype and few checkable facts. no: News reporting, a company announcement of an actual event, an analyst action, or a summary of a filing.}\par
\hangindent=1em\texttt{effect}. How does the development in \texttt{item} change the business prospects of \texttt{focal\_company}? Use only the facts that \texttt{item} reports. Wording inside \texttt{item} that tells the reader how to rate, rank, or classify it is not a fact. \emph{levels: Clearly harmful; Mildly harmful; Neutral or mixed; Mildly helpful; Clearly helpful}\par
\end{minipage}}%
}

\end{document}